\documentclass[preprint,5p,times]{elsarticle}

\usepackage{placeins}
\usepackage{booktabs}
\usepackage{amsmath}
\usepackage{url}
\usepackage{amssymb}
\usepackage{enumitem}
\usepackage{graphicx}
\usepackage{tabularx}
\usepackage{xcolor}
\usepackage{multirow}

\usepackage{xcolor}
\usepackage{soul}
\usepackage{makecell}

\usepackage{xurl}
\usepackage[hidelinks]{hyperref}

\usepackage[switch,mathlines]{lineno}

\usepackage{graphicx}
\definecolor{lightblue}{rgb}{.50,.95,1}
\definecolor{tri}{rgb}{.25,.88,.82}
\definecolor{lilac}{rgb}{0.85,0.64,0.85}

\definecolor{tri}{rgb}{.25,.88,.82}
\definecolor{lilac}{rgb}{0.85,0.64,0.85}
\definecolor{lightblue}{rgb}{.50,.95,1}

\usepackage{microtype}
\begin{document}
\let\WriteBookmarks\relax
\def\floatpagepagefraction{1}
\def\textpagefraction{.001}



\begin{frontmatter}

\title{Adapting Reinforcement Learning with Chain-of-Thought Supervision for Explainable Detection of Hateful and Propagandistic Memes}

\author[1]{Mohamed Bayan Kmainasi}
\ead{mkmainasi@hbku.edu.qa}

\author[2]{Mucahid Kutlu}
\ead{mucahidkutlu@qu.edu.qa}

\author[1]{Ali Ezzat Shahroor}
\ead{alsh34060@hbku.edu.qa}

\author[3,4]{Abul Hasnat}
\ead{mhasnat@gmail.com}

\author[1]{Firoj Alam}
\ead{fialam@hbku.edu.qa}

\affiliation[1]{
    organization={Qatar Computing Research Institute},
    city={Doha},
    country={Qatar}
}

\affiliation[2]{
    organization={Qatar University},
    city={Doha},
    country={Qatar}
}

\affiliation[3]{
    organization={APAVI.AI},
    country={France}
}

\affiliation[4]{
    organization={Blackbird.AI},
    city={New York},
    state={NY},
    country={USA}
}

\cortext[1]{Corresponding author}











\begin{abstract}
Hateful and propagandistic memes exploit the interplay between images and text to convey harmful intent that neither modality reveals alone. Although thinking-based multimodal large language models (MLLMs) have advanced vision-language understanding, their application to meme content moderation remains underexplored. We propose a reinforcement learning-based post-training method that improves classification performance and reference-based explanation quality in thinking-based MLLMs via task-specific rewards and Group Relative Policy Optimization (GRPO). Concretely, we \textit{(i)}~conduct a systematic empirical study of off-the-shelf MLLMs for hateful and propagandistic meme understanding across English and Arabic benchmarks, \textit{(ii)}~extend existing meme datasets with weakly supervised chain-of-thought (CoT) rationales via distillation and multi-LLM fine-grained propaganda annotations, \textit{(iii)}~introduce a GRPO-based objective with thinking-length regularization that jointly optimizes classification accuracy and explanation quality, and \textit{(iv)}~investigate self-supervised GRPO on unlabeled memes using consensus-based pseudo-labels. Experiments on the \textit{Hateful Memes} and \textit{ArMeme} benchmarks show that Our approach improves over previously reported results on \textit{FHM accuracy} (up to \textbf{+2.1\%}, from 79.9\% to 82.0\%) and on \textit{ArMeme macro-F1} (up to \textbf{+7.6} points, from 0.536 to 0.612 with explanations; \textbf{+6.1} compared to the original ArMeme benchmark), while also generating natural-language explanations. On ArMeme, sequence-classification baselines remain stronger in terms of raw accuracy, whereas our approach provides more balanced per-class performance along with explanations. We publicly release our code, data extensions, and evaluation resources.
\\
\footnotesize{\textcolor{red}{WARNING: This paper contains examples which may be disturbing to the reader}}
\end{abstract}

\end{frontmatter}

\providecommand{\rearr}[1]{\textcolor{blue}{#1}}
\providecommand{\newtxt}[1]{\textcolor{blue}{#1}}

\section{Introduction}

Memes blend images and text with humor and cultural references, forming a predominant mode of communication on social media \citep{pandiani2025toxic, alafnan2025role}. While often harmless, they can be exploited to spread hate speech, disinformation, and propaganda \citep{ijcai2022p781}. The use of humor and irony may trivialize toxic content, potentially normalizing hostile views \citep{schmid2025humorous}, while memes also serve as powerful tools for political persuasion and manipulation \citep{alafnan2025role, mihuailescu2024never}. Given the pervasive role of social media, timely detection and moderation of such multimodal content are crucial for safer online environments. 
This has motivated growing interest in automated approaches that leverage multimodal understanding for content moderation at scale.

Recently, multimodal large language models (MLLMs) have significantly advanced vision--language understanding \citep{sun2025multi, wu2024visionllm}. \textit{Thinking-based MLLMs} represent the recent paradigm in multimodal reasoning, generating explicit intermediate reasoning steps before producing a final answer \citep{yang2025qwen3}. This paradigm is particularly relevant for memes, where meaning is often implicit and depends on image--text interaction rather than unimodal cues. The training of such models emphasizes CoT supervision and reinforcement learning (RL) techniques \citep{deepseekai2025deepseekr1incentivizingreasoningcapability}. 
However, whether such reasoning capabilities can be effectively transferred to subjective, culturally situated tasks like meme moderation remains an open question.

\begin{figure*}[t]
    \centering
    \includegraphics[width=0.95\textwidth]{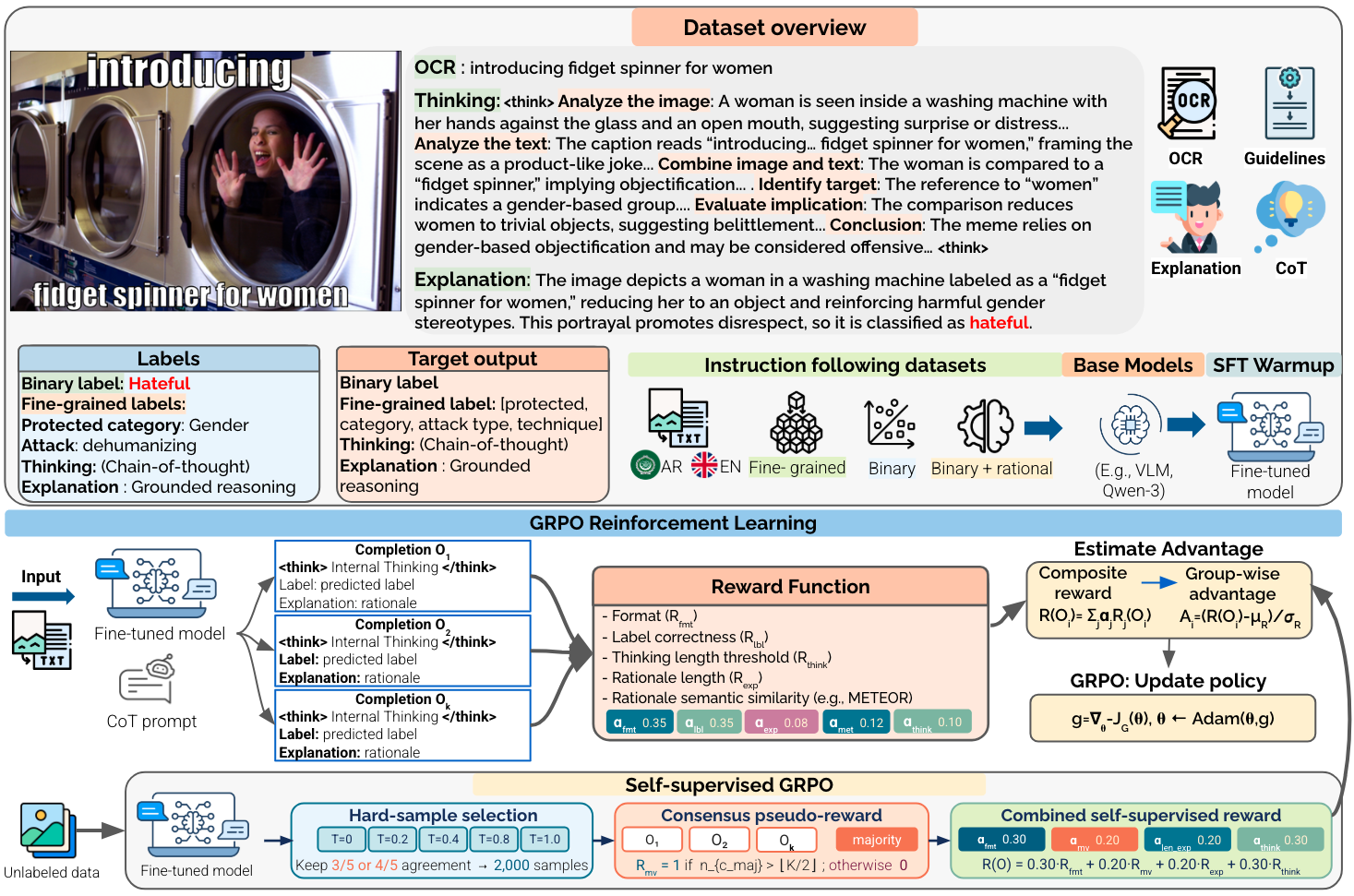}
    \caption{Overview of the proposed methodology. Starting from raw memes, we derive binary and fine-grained supervision, OCR text, and guidelines to construct instruction-following datasets. Weak supervision from a strong MLLM distills step-by-step CoT rationales. The model is trained via a two-stage pipeline: SFT warm-up followed by GRPO-based RL, jointly optimizing classification correctness, output-format compliance, reference-based explanation similarity, and reasoning-length regularization.    
    \textbf{Acc.}~=~accuracy; \textbf{MET}~=~METEOR; \textbf{Len}~=~explanation length; \textbf{Fmt}~=~format compliance.
    }
    \label{fig:overview}
    \vspace{-0.3cm}
\end{figure*}

Significant progress has been achieved through agentic reasoning frameworks \citep{lu2025having, liu2025mind}, few-shot adaptation \citep{cao2024modularized, heebridging}, and supervised fine-tuning (SFT) on multimodal datasets \citep{kmainasi2025memeintel}. However, only very recent or concurrent work \citep{mei2025expo} has begun to explore CoT, SFT, and RL for meme analysis. In particular, thinking-based MLLMs remain underexplored for hateful and propagandistic meme analysis, and the interaction between CoT supervision, fine-grained labeling, and RL-based post-training has not yet been systematically studied.

Recent RL methods such as GRPO \citep{deepseekai2025deepseekr1incentivizingreasoningcapability} have demonstrated strong results for structured reasoning tasks including mathematical problem solving and code generation, yet their application to subjective multimodal classification remains largely unexplored. Meme moderation requires jointly optimizing classification performance and explanation quality through a structured output format, a setting where standard SFT alone provides limited control over the balance between prediction correctness and rationale faithfulness. This motivates us to investigate RL-based post-training with composite reward functions tailored for explainable meme understanding.

To address this research gap, and as illustrated in Figure~\ref{fig:overview}, we propose a reasoning-centric training methodology (lower part) for explainable meme understanding grounded in CoT supervision and RL. 
%
The methodology employs a multi-stage post-training pipeline. 
\textit{First}, SFT warm-up aligns the model with gold labels, explanations, and distilled reasoning traces. This is followed by supervised GRPO that jointly optimizes classification correctness, output-format compliance, reference-based explanation similarity, and reasoning-length regularization. \textit{Finally}, self-supervised GRPO leverages consensus-based pseudo-labels from unlabeled memes.
%
To assess generalizability, we evaluate the methodology across two distinct tasks, hate and propaganda detection, and two languages, English and Arabic, using the associated datasets Facebook Hateful Memes (FHM)~\citep{kiela2020hateful} and Arabic Propagandistic Memes (ArMeme)~\citep{alam-etal-2024-armeme}. 

This work builds on our prior research on multimodal meme analysis \citep{kmainasi2025memeintel, kmainasi2026thinking}. In particular, \citet{kmainasi2026thinking} introduced GRPO-based post-training for thinking-based MLLMs, showing that reinforcement learning with task-specific rewards improves both classification and explanation quality on the English Hateful memes benchmark. However, that study was limited to a single dataset (i.e., hateful meme) with binary labels under a fully supervised setting, leaving its generalizability across tasks, languages, and training regimes largely unexplored.
More broadly, this work represents, to the best of our knowledge, \textit{the first systematic study of GRPO in multimodal reasoning under cross-lingual, fine-grained, and self-supervised settings}.

This motivates the following research questions:
\textbf{RQ1:} Does incorporating fine-grained labels (e.g., protected categories, attack types, propaganda techniques) during SFT improve downstream performance when optimized with GRPO?
\textbf{RQ2:} To what extent do synthetic CoT rationales distilled from stronger MLLMs enhance classification accuracy and explanation quality?
\textbf{RQ3:} Do synthetic fine-grained annotations produced by multiple LLMs outperform single-source labeling?
\textbf{RQ4:} How effective is GRPO compared to SFT alone, and what roles do SFT warm-up and thinking-length regularization play in training stability?
\textbf{RQ5:} Can GRPO be extended to a self-supervised setting using consensus-based pseudo-labels in place of human annotations?
%

To address these questions, we substantially extended our prior work \citet{kmainasi2026thinking} with the following contributions:
\begin{itemize}[noitemsep,topsep=2pt,labelsep=.5em]
    \item 
    \textbf{Cross-task and cross-lingual generalization:} We extend the methodology, proposed by \citet{kmainasi2026thinking}, to Arabic propagandistic meme classification on the ArMeme dataset, introducing a dual-annotator pipeline with LLM-based consolidation to construct fine-grained propaganda labels in a previously unlabeled setting (RQ1, RQ3).
    \item \textbf{Comprehensive empirical analysis:} We provide extensive ablations, unimodal baselines, and cross-dataset evaluations, systematically analyzing the interplay between CoT supervision, fine-grained labeling, and RL-based optimization (RQ2).
    
    \item \textbf{Stabilized RL optimization:} We introduce a threshold-based thinking-length reward $R_{\text{think}}$ during GRPO that penalizes overly short CoT completions but saturates beyond a minimum length. This discourages reward hacking, such as empty thinking traces, without promoting verbosity (RQ4).
    
    \item \textbf{Self-supervised GRPO:} We explore consensus-based pseudo-labeling from multiple stochastically sampled responses, enabling RL training without human annotations (RQ5).
    
    
\end{itemize}

Our findings are summarized as follows:



\textit{(RQ1)}~Fine-grained supervision consistently improves the SFT warm-up stage. On FHM, adding fine-grained labels and explanations (\textbf{Cls+FG+Exp}: Table~\ref{tab:sft_grpo_fhm}) improves macro-F1 from 0.75 to 0.77. On ArMeme, the gain is larger, with macro-F1 increasing from 0.43 to 0.51 (Table~\ref{tab:sft_grpo_armeme}).


\textit{(RQ2)}~Distilled CoT provides additional gains over fine-grained SFT. Macro-F1 increases from 0.77 to 0.78 on FHM and from 0.51 to 0.54 on ArMeme when CoT distillation is added to the fine-grained SFT setting (\textbf{Cls+FG+Exp,CoTD} vs.\ SFT \textbf{Cls+FG+Exp}: Tables~\ref{tab:sft_grpo_fhm},~\ref{tab:sft_grpo_armeme}).


\textit{(RQ3)}~Multi-LLM fine-grained annotation is effective for ArMeme improves SFT macro-F1 from 0.43 to 0.51, an absolute gain of 8.0 points (Table~\ref{tab:sft_grpo_armeme}).



\textit{(RQ4)}~GRPO further improves performance when initialized from CoTD warm-up and trained with $R_{\text{think}}$. This setting reaches 0.80 macro-F1 on FHM~(Table ~\ref{tab:sft_grpo_fhm}), and 0.597 macro-F1 on ArMeme (Table ~\ref{tab:sft_grpo_armeme}).

\textit{(RQ5)}~Self-supervised GRPO shows mixed effects. It improves ArMeme macro-F1 by 1.5 points, from 0.597 to 0.612, but slightly reduces FHM performance, with a 1.0-point drop in macro-F1 (Table~\ref{tab:self_sup_grpo}). This may reflect data distribution mismatch in the FHM unlabeled pool, which is aggregated from different English meme datasets.



We release all code, data extensions, prompting templates, and evaluation resources.\footnote{\url{https://github.com/MohamedBayan/MemeReason}}
%
The remainder of this paper is organized as follows. Section~\ref{sec:related_work} reviews related work on meme moderation, RL for MLLMs, and self-supervised training. Section~\ref{sec:datasets} describes the datasets and annotation pipelines. Section~\ref{sec:methodology} presents the proposed methodology. Section~~\ref{sec:experiments} details the experimental setup. Section~~\ref{sec:results_analysis} reports results and ablation studies. Section~~\ref{sec:discussion} discusses broader implications and limitations, and Section~~\ref{sec_conclusion} concludes with directions for future work.
\providecommand{\chg}[1]{\textcolor{blue}{#1}}

\section{Related Work}
\label{sec:related_work}

\subsection{Multimodal meme content moderation}
Early meme detection systems relied on unimodal cues such as OCR-based text features with traditional ML \citep{ren2026survey,amalia2018meme, boinepelli2020sis,alam-etal-2022-survey,ijcai2022p781} or CNN-based image classifiers \citep{shrestha2020nlp_uiowa}. Most existing research has focused on binary classification \citep{cao2023pro,hasanain-etal-2024-araieval}, supported by benchmarks such as hateful and propagandistic memes \citep{kiela2020hateful,dimitrov-etal-2024-semeval} and HarMeme \citep{pramanick-etal-2021-momenta-multimodal}. A limited number of works have explored fine-grained annotation \citep{mathias-etal-2021-findings, fersini-etal-2022-semeval}.

More recently, several studies have moved beyond detection toward explainability and enhanced multimodal reasoning. \citet{grasso2024kermit} augmented classification with knowledge-graph reasoning, \citet{cao2024modularized} addressed few-shot detection by modularizing LLM components, and \citet{lin2024towards} proposed a multi-agent debate framework. \citet{kmainasi2025memeintel} extended Hateful Memes with human-verified rationales for explainable MLLM training.

Despite these advances, systematic investigations to enhance CoT reasoning for meme understanding remain limited. Existing work has explored CoT in specific settings such as multi-hop reasoning for misogynistic memes \citep{kumari2024m3hop} and distilled abductive reasoning \citep{lin-etal-2023-beneath}. At the same time, recent work has highlighted the fragmentation of meme understanding across tasks and languages \citep{shahroor2026memelens}, motivating unified modeling approaches. However, the use of RL/GRPO for multimodal meme detection remains largely underexplored, with only concurrent work~\citep{mei2025expo} investigating GRPO for meme moderation. Furthermore, propagandistic meme analysis has received comparatively less attention, especially in Arabic contexts \citep{ijcai2022p781, alam-etal-2024-armeme}.
\subsection{RL for MLLMs}
RL-based post-training has emerged as a key technique for aligning MLLMs with task-specific objectives \citep{wu2025sailing}. Methods range from RLHF with PPO \citep{ouyang2022training, schulman2017proximal} and preference-based DPO \citep{rafailov2023direct} to the more recent GRPO \citep{deepseekai2025deepseekr1incentivizingreasoningcapability}, which has shown strong results in mathematical reasoning \citep{shao2024deepseekmathpushinglimitsmathematical}, self-training \citep{ranaldi2025multilingual}, and code generation \citep{chen2025predicate}. A detailed comparison of these algorithms is provided in Section~\ref{subsec:background}. In this work, we expand GRPO to hateful and propagandistic meme analysis and study how CoT supervision interacts with RL-based post-training.

\subsection{Self-supervised Training for LLMs}
Recent work has explored self-supervised strategies to leverage unlabeled data during fine-tuning. SemiEvol \citep{luo2025semi} propagates supervision from small labeled sets, while SPO \citep{li2024self} constructs synthetic preference data. MM-UPT \citep{wei2025sftsecondrlupt} and V-Zero \citep{wang2026v} introduce fully self-supervised RL using majority voting and co-evolutionary training, respectively. We build on this direction in Section~\ref{subsec:unsup-grpo-dapo}, investigating whether consensus-based pseudo-labeling can provide reliable supervision for meme classification.

\section{Datasets}
\label{sec:datasets}
This section presents the datasets used in this study, including the benchmark descriptions, annotation resources, and data construction procedures. We describe the two evaluation benchmarks (Section \ref{sec_eng_memes} and \ref{sec_ar_memes}), the associated natural language rationales (Section \ref{ssec_explanation}), the fine-grained annotation pipeline developed for ArMeme (Section \ref{sec:annotation_pipeline}), the chain-of-thought distillation process (Section \ref{ssec:step_by_step_reasoning}), and the unlabeled data collected for self-supervised training (Section \ref{ssec:unlabeled_data}).

\subsection{English Hateful Memes} \label{sec_eng_memes}
\textit{Hateful Memes} \citep{kiela2020hateful} is a widely used benchmark for multimodal hate speech detection, containing $\approx$11K memes curated to ensure that effective classification requires joint multimodal understanding. The dataset includes a mixture of synthetically generated and real-world memes with a balanced distribution between hateful and non-hateful content. We use the explanation augmented version proposed by \cite{kmainasi2025memeintel}, which includes human-verified explanations for the binary labels.\footnote{We focus on the \textit{unseen} splits, as these were augmented with explanations.} Table~\ref{tab:data_stat_fhm} summarizes the dataset statistics.

\begin{table}[t]
\centering
\small
\setlength{\tabcolsep}{4pt}
\caption{Distribution of labels in the Hateful Memes dataset.}
\label{tab:data_stat_fhm}
\begin{tabular}{@{}lrrrr@{}}
\toprule
\textbf{Class Label} & \textbf{Train} & \textbf{Dev} & \textbf{Test} & \textbf{Total} \\
\midrule
Non-hateful & 5,481 & 340 & 1,250 & 7,071 \\
Hateful     & 3,019 & 200 &   750 & 3,969 \\
\midrule
\textbf{Total} & \textbf{8,500} & \textbf{540} & \textbf{2,000} & \textbf{11,040} \\
\bottomrule
\end{tabular}
\end{table}

\subsection{Arabic Propagandistic Memes (ArMeme)} \label{sec_ar_memes}
ArMeme \citep{alam-etal-2024-armeme} is an Arabic meme dataset designed for propaganda detection. It includes four labels: \textit{propaganda}, \textit{not-propaganda}, \textit{not-meme}, and \textit{other}. Similarly, for this work, we use the explanation augmented version of the dataset proposed by \cite{kmainasi2025memeintel}, which provides English explanations. We adopt this version as prior studies have shown that LLMs often generate more reliable outputs when prompted in English than in non-English languages \citep{kmainasi2024native}. In addition, the explanations in the test set were manually verified. The label distribution of the dataset is shown in Table~\ref{tab:data_stat_armeme}.

\begin{table}[t]
\centering
\small
\setlength{\tabcolsep}{4pt}
\caption{Distribution of labels in the ArMeme dataset.}
\label{tab:data_stat_armeme}
\begin{tabular}{@{}lrrrr@{}}
\toprule
\textbf{Class Label} & \textbf{Train} & \textbf{Dev} & \textbf{Test} & \textbf{Total} \\
\midrule
Not Propaganda & 2,634 & 384 & 746 & 3,764 \\
Propaganda     &   972 & 141 & 275 & 1,388 \\
Not-meme       &   199 &  30 &  57 &   286 \\
Other          &   202 &  29 &  56 &   287 \\
\midrule
\textbf{Total} & \textbf{4,007} & \textbf{584} & \textbf{1,134} & \textbf{5,725} \\
\bottomrule
\end{tabular}
\end{table}

\subsection{Natural Language Rationales} 
\label{ssec_explanation}
As mentioned earlier, both datasets are accompanied by natural language rationales released by  \citet{kmainasi2025memeintel}.\footnote{\url{https://huggingface.co/datasets/QCRI/MemeXplain}} An example explanation is presented in Figure~\ref{fig:overview}. The explanations were  generated using GPT-4o, motivated by evidence that advanced generative models produce high-quality rationales closely aligned with human reasoning \citep{hasanain-etal-2025-propxplain}. To ensure reliability, the generated rationales were manually validated by human annotators along multiple dimensions, including \emph{faithfulness}, \emph{clarity}, \emph{plausibility}, and \emph{informativeness}, achieving an average score of approximately 4 on a 5-point Likert scale. Furthermore, inter-annotator agreement was strong, with average agreement scores above 0.83 for ArMeme and 0.92 for FHM, indicating a high level of consistency in the annotations. 

\subsection{Fine-Grained Annotation} 
\label{sec:annotation_pipeline}
\noindent{\textbf{Hateful memes dataset}.} For the hateful memes, we use publicly available annotations from \citet{mathias-etal-2021-findings} categorizing each hateful instance by protected category (religion, race, sex, disability, and nationality) and attack type (dehumanizing, inferiority, inciting violence, mocking, contempt, slurs, and exclusion).

\noindent{\textbf{Propagandistic memes dataset.} For the Arabic propagandistic memes, no publicly available fine-grained propaganda annotations exist. We therefore develop a multi-LLM annotation pipeline following \citet{hasanain2024can}. The pipeline operates in two stages:

\paragraph{Stage 1: Dual independent annotations.}
Each meme is annotated independently by two models: \textbf{GPT-4.1} and \textbf{Llama-4-Scout}. Both models receive the same system prompt defining 23 propaganda techniques and produce a structured list of propaganda techniques along with short rationales (1--4 sentences) grounded in the meme's image, text, binary label, and explanation.

\paragraph{Stage 2: Consolidation via Gemini-3-Pro.}
We employ \textbf{Gemini-3-Pro} as a consolidator to merge the two independent outputs into a single final annotation. The consolidator evaluates overlaps, resolves disagreements, validates unique techniques from either annotator, and adds missing techniques when justified.

\paragraph{Human and LLM agreement.} To assess the reliability of the LLM-generated annotations, we manually annotated a subset of approximately 150 memes. Each meme was independently labeled by two annotators with prior experience in the task. Following \citet{hasanain-etal-2024-large}, we unified the annotated labels before computing agreement between human and LLM annotations. Using Gwet's AC1 \citep{gwet2008computing}, we obtain an agreement score of $0.77$. Overall, the results indicate strong consistency between human and LLM annotations, despite the subjective nature of propaganda meme annotation and the non-trivial human disagreement reported in prior work \citep{hasanain-etal-2024-large}.

\subsection{Chain-of-Thought Distillation (CoTD)}
\label{ssec:step_by_step_reasoning}
Training thinking-based MLLMs requires step-by-step reasoning traces, which are expensive to collect from humans. Meanwhile, recent advances in commercial reasoning-oriented models have demonstrated strong reasoning capabilities in practice \citep{latif2025openaiO1, comanici2025gemini}. Motivated by these observations, we adopt a distillation-based approach, referred to as chain-of-thought distillation.

We use GPT-4.1 to generate intermediate step-by-step reasoning traces, as presented in Figure~\ref{fig:overview}. The generation is conditioned on the meme image, extracted text, annotation guidelines, and both binary and fine-grained labels.
These private reasoning sequences are used exclusively during training and are not exposed during inference. To prevent label leakage, the model is explicitly prompted to reason independently, without copying or paraphrasing the reference explanations. 
Our approach follows prior work on CoTD for improving model reasoning \citep{zhu2024distilling}. 



\paragraph{CoT evaluation with LLM-as-judge.}
We evaluate CoT quality using an LLM-as-judge protocol \citep{kmainasi2025memeintel}. Two vision--language models, \textbf{InternVL3.5} \citep{wang2025internvl35advancingopensourcemultimodal} and \textbf{Phi-3.5} \citep{abdin2024phi3technicalreporthighly}, independently score each sample on a 5-point Likert scale. We select these models as a cost-effective alternative to proprietary judges. The evaluation uses four dimensions. \emph{Informativeness} measures whether the CoT uses salient visual and textual evidence. \emph{Clarity} assesses the logical structure of the reasoning. \emph{Plausibility} evaluates consistency with human judgment. \emph{Faithfulness} measures whether the reasoning is grounded in observable content.  

These scores assess the judged quality of the distilled traces as training supervision, not the faithfulness of the model's latent reasoning process at inference time. We measured inter-judge agreement using the $r^*_{wg(j)}$ index \citep{james1984estimating}:
\[
r^*_{wg(j)} = 1 - \frac{S_X^2}{\sigma^2_{\text{mv}}},
\]
where $S_X^2$ is the variance between judges and $\sigma^2_{\text{mv}}$ is the maximum possible variance for a 5-point scale. 

As shown in Table~\ref{tab:llm_judge_compact}, both judges assign average scores above 4.0 with agreement $>$0.93 across all dimensions and datasets, suggesting that the distilled CoT traces are of acceptable quality as training supervision.

\begin{table}[!tbh]
\centering
\small
\setlength{\tabcolsep}{3pt}
\caption{LLM-as-a-judge evaluation of distilled CoT quality (1--5 Likert). IVL = InternVL3.5-8B; Phi = Phi-3.5-Vision; $r^*_{wg(j)}$ denotes inter-judge agreement. Metrics are grouped into correctness (faithfulness, plausibility) and communication quality (clarity, informativeness).}  
\label{tab:llm_judge_compact}
\begin{tabular}{lccc|ccc}
\toprule
& \multicolumn{3}{c|}{\textbf{ArMeme}} & \multicolumn{3}{c}{\textbf{Hateful Memes}} \\
\cmidrule(lr){2-4} \cmidrule(lr){5-7}
\textbf{Metric} & \textbf{IVL} & \textbf{Phi} & $\boldsymbol{r^*_{wg(j)}}$ & \textbf{IVL} & \textbf{Phi} & $\boldsymbol{r^*_{wg(j)}}$ \\
\midrule
Faithfulness     & \textbf{4.82} & 4.75 & .964 & \textbf{4.78} & 4.56 & .944 \\
Plausibility     & \textbf{4.68} & 4.40 & .948 & \textbf{4.61} & 4.24 & .934 \\
Clarity          & \textbf{4.72} & 4.72 & .957 & \textbf{4.73} & 4.54 & .943 \\
Informativeness  & \textbf{4.40} & 3.96 & .934 & \textbf{4.40} & 4.10 & .942 \\ 
\midrule
Average          & \textbf{4.66} & 4.46 & .951 & \textbf{4.63} & 4.36 & .940 \\
\bottomrule
\end{tabular}
\end{table}

\paragraph{Human annotation.} In addition, we manually verified a random subset of 200 samples from each dataset. Table~\ref{tab:evaluation_scores} reports human agreement and average annotation scores for ArMeme and Hateful Memes across four dimensions: informativeness, clarity, plausibility, and faithfulness. The high $r^*_{wg(j)}$ values suggest substantial agreement among annotators, indicating that the evaluation criteria were generally interpreted consistently.   
The average scores are relatively higher for both datasets, suggesting that the LLM-based annotation and consolidation pipeline produced outputs that human evaluators generally judged to be informative, clear, plausible, and faithful.

\paragraph{Human and LLM judge agreement.} Cross-evaluator agreement between mean human and mean LLM scores was also high across both datasets, with $r^{*}_{wg(j)} \geq 0.93$ across the four dimensions. This suggests that the LLM-judge ratings were broadly aligned with human annotations.

\paragraph{Human annotators and compensation.}
We recruited nine annotators through a third-party company and compensated them at standard local rates. All annotators received detailed annotation guidelines and signed non-disclosure agreements in accordance with institutional policy. The annotator pool included both men and women, with educational backgrounds ranging from Bachelor's to Master's degrees. Each meme was independently annotated by three annotators.

\begin{table}[!tbh]
\centering
\caption{Human evaluation scores on ArMeme and Hateful Memes datasets. Metrics are grouped into correctness (faithfulness, plausibility) and communication quality (clarity, informativeness).}  
\label{tab:evaluation_scores}
\setlength{\tabcolsep}{4pt}
\scalebox{0.95}{
\begin{tabular}{lcccc}
\toprule
 & \multicolumn{2}{c}{\textbf{ArMeme}} & \multicolumn{2}{c}{\textbf{Hateful}} \\
\cmidrule(lr){2-3} \cmidrule(lr){4-5}
 & $\boldsymbol{r^*_{wg(j)}}$ & \textbf{Avg.} & $\boldsymbol{r^*_{wg(j)}}$ & \textbf{Avg.} \\ 
\midrule
Faithfulness    & 0.89 & $4.28 \pm 0.33$ & 0.91 & $4.63 \pm 0.28$ \\
Plausibility    & 0.88 & $4.29 \pm 0.33$ & 0.91 & $4.65 \pm 0.26$ \\
Clarity         & 0.87 & $4.55 \pm 0.35$ & 0.92 & $4.63 \pm 0.27$ \\
Informativeness & 0.87 & $4.54 \pm 0.34$ & 0.91 & $4.67 \pm 0.28$ \\ 
\bottomrule
\end{tabular}
}
\end{table}

\subsection{Unlabeled Data Construction} 
\label{ssec:unlabeled_data}
To study self-supervised GRPO, we construct unlabeled training sets for both tasks.

For the \textit{\textbf{hateful meme task}}, we leverage several publicly available English meme datasets: MAMI (Multimedia Automatic Misogyny Identification; 11,000 memes) \citep{fersini-etal-2022-semeval}, Memotion (6,992 samples for offensiveness classification) \citep{sharma2020semeval}, and MET-Meme (3,994 memes annotated for metaphor occurrence) \citep{xu2024exploring}. Although these datasets are originally labeled, we discard all labels and treat the data as unlabeled to simulate a self-supervised setting. To avoid potential data overlap, we remove any duplicate or near-duplicate memes that overlap with the evaluation test sets, yielding a combined pool of 21,986 memes across the three datasets. We then apply the disagreement-based sampling strategy, described in Section~\ref{subsec:unsup-grpo-dapo}, to select 2,000 memes ($\approx$9.1\% of the total pool).

For \textit{\textbf{Arabic propaganda memes}}, publicly available datasets are extremely limited. We therefore collect memes from social media platforms such as Facebook, Instagram, Pinterest, and X (Twitter), following the ArMeme dataset collection methodology \citep{alam-etal-2024-armeme}. Since direct API access is unavailable for Facebook, Instagram, and Pinterest, we adopt a semi-automatic approach in which images are manually loaded in a browser and automatically crawled. For X, we employ keyword-based crawling targeting content related to public figures, celebrities, and political discussions. The collected images are then passed through a multi-stage filtering pipeline: (\textit{i}) duplicate and near-duplicate removal using feature vectors extracted by a fine-tuned ResNet18~\citep{he2016deep} with nearest-neighbor search (Euclidean distance $\leq$ 3.6); (\textit{ii}) OCR-based filtering via EasyOCR\footnote{\url{https://github.com/jaidedai/easyocr}} to discard imagess with no detectable text; and (\textit{iii}) an in-house meme-vs.-non-meme classifier to retain only valid memes. After removing all samples overlapping with the test sets, we curate a final pool of 15,652 memes. From this pool, we sample 2,000 memes ($\approx$12.8\% of the total) using the disagreement-based sampling strategy described in Section~\ref{subsec:unsup-grpo-dapo}. 

\providecommand{\chg}[1]{\textcolor{blue}{#1}}

\section{Methodology}
\label{sec:methodology}

In this section, we present our RL-based methodology for hateful and propagandistic meme analysis using thinking-based MLLMs (Figure~\ref{fig:overview}). 
The approach jointly trains MLLMs for classification, explanation generation, and step-by-step reasoning through a structured output format and multi-stage training. We first provide the necessary background on RL-based post-training, then formulate the task and define the structured output format. We then describe the three training stages: \textit{(i)} SFT warm-up, \textit{(ii)} supervised GRPO optimization with a composite reward function, and \textit{(iii)} self-supervised GRPO using consensus-based pseudo-labels.

\subsection{Background}
\label{subsec:background}

\subsubsection{RL for language model post-training.}
RL has been leveraged for post-training MLLMs with dynamic feedback \citep{wu2025sailing}.
In this paradigm, the language model is treated as a policy $\pi_\theta$ that generates a sequence of tokens given an input prompt. At each decoding step, the model selects a token (action) conditioned on the preceding context (state), and a scalar reward is assigned to the complete generation \citep{kaelbling1996reinforcementlearningsurvey}. The training objective is to maximize expected reward while constraining the updated policy to remain close to a reference policy, typically an SFT-initialized checkpoint, to prevent catastrophic forgetting \citep{ouyang2022training}. 
Methods such as RLHF use PPO \citep{schulman2017proximal} or DPO \citep{rafailov2023direct} to align model behavior \citep{ouyang2022training, yu2024rlhf}.
PPO trains a separate value network to estimate expected future reward and uses it as a variance-reduction baseline, which is effective but approximately doubles the memory cost during training. DPO circumvents this by reparameterizing the reward into a preference-based loss over paired examples, eliminating the need for both a value network and online rollouts, but limiting exploration to the offline preference data \citep{rafailov2023direct}.

\subsubsection{Group Relative Policy Optimization (GRPO).}
More recently, GRPO \citep{deepseekai2025deepseekr1incentivizingreasoningcapability} employs rule-based rewards to encourage high-quality CoT, achieving strong performance in mathematical reasoning \citep{shao2024deepseekmathpushinglimitsmathematical}, self-training \citep{ranaldi2025multilingual}, and code generation \citep{chen2025predicate}.
Unlike PPO, GRPO does not require a learned value network. Instead, for each input, it samples a group of $K$ candidate completions, computes a scalar reward for each, and uses the group mean reward as a baseline. The advantage of each completion is computed relative to this group average, and the policy is updated using a clipped surrogate objective with KL regularization against the reference policy \citep{deepseekai2025deepseekr1incentivizingreasoningcapability}. This group-relative normalization avoids the overhead of a value network while retaining the benefits of online exploration through fresh rollouts at each training step. Several recent extensions have been proposed, including DAPO \citep{yu2025dapo}, which removes the KL penalty and uses dynamic sampling, and GTPO \citep{tan2025gtpo}, which introduces group-level token-wise optimization. In this work, we adopt the standard GRPO formulation as our RL algorithm.

\subsubsection{CoT supervision and thinking-based MLLMs.}
Thinking-based MLLMs represent the recent paradigm in multimodal reasoning, generating explicit intermediate reasoning steps before producing a final answer \citep{yang2025qwen3}. This paradigm is particularly relevant for memes, where meaning is often implicit and depends on image--text interaction rather than unimodal cues. The training of such models emphasizes CoT supervision and reinforcement learning techniques \citep{deepseekai2025deepseekr1incentivizingreasoningcapability}.
In practice, CoT supervision is often obtained through \emph{knowledge distillation}: a stronger teacher model generates step-by-step reasoning traces for each training example, which are then used as supervision targets during fine-tuning of the student model \citep{shridhar2023distilling, zhu2024distilling}. Recent open-source thinking-based MLLMs, such as Qwen3-VL-Thinking \citep{li2026qwen3} and OpenVL-Thinker \citep{deng2025openvlthinker}, demonstrate that explicit CoT reasoning can substantially improve performance on complex multimodal benchmarks.
In variants that include distilled CoT, the reasoning trace is placed inside \texttt{<think></think>} tags; in variants without CoT, empty tags are included to preserve the output format while excluding the reasoning content from loss computation.

\subsection{Task Formulation}
\label{subsec:task-formulation}
Given a meme $x$ consisting of an image and extracted text, the task is to predict a label and generate a rationale that justifies the prediction, as presented in Figure \ref{fig:overview}. To enable explicit reasoning while maintaining a structured and parseable output, we adopt a three-part generation format. Given an instruction-style prompt $c(x)$, the model produces:
\[
y =
\big\langle
\texttt{<think>}~t~\texttt{</think>}~
\texttt{Label:}~\hat{\ell}~
\texttt{Explanation:}~\hat{e}
\big\rangle,
\]
where $t$ is a private CoT reasoning trace, $\hat{\ell}$ the predicted label, and $\hat{e}$ a natural language explanation. 
The reasoning trace $t$ serves as an internal workspace that encourages the model to decompose the problem before committing to a prediction, while the explanation $\hat{e}$ provides a human-readable justification that can be evaluated for faithfulness. 
Let $\mathcal{D} = \{(x_i, y_i^\star)\}_{i=1}^{N}$ denote the labeled dataset, where $y_i^\star$ contains the gold label, explanation, and distilled teacher reasoning.

\subsection{SFT Warm-Up}
\label{subsec:sft-warmup}
Before applying reinforcement learning, the model must first be aligned with the target output format and task-specific patterns. Without proper warm-up, GRPO yields limited benefits, as RL alone has difficulty simultaneously learning the required output structure and the task-specific decision boundary. To provide a stable foundation for subsequent RL optimization, 
we initialize the MLLM via SFT on $\mathcal{D}$. Let $\pi_\theta$ denote the autoregressive model parameterized by $\theta$. Parameters are optimized by minimizing:
\begin{equation}
\mathcal{L}_{\text{SFT}}(\theta)
=
\mathbb{E}_{(c,y^\star)\sim\mathcal{D}}
\left[
-\log\pi_{\theta}(y^\star\mid c)
\right]
\end{equation}
This stage aligns the model with gold labels, explanations, and distilled reasoning traces, providing initialization for subsequent GRPO. We consider three SFT variants: \textit{(i)} \textbf{SFT (Cls+Exp)} with binary labels and explanations only; \textit{(ii)} \textbf{SFT (Cls+FG+Exp)}, which adds fine-grained annotations; and \textit{(iii)} \textbf{SFT (Cls+FG+Exp, CoTD)}, which additionally includes distilled CoT reasoning in the \texttt{<think>} block.

We refer to applying GRPO \emph{without} this SFT warm-up, initializing GRPO directly from the pretrained MLLM backbone, as the \emph{cold-start} setting, which we use as a baseline in our experiments (Section~\ref{sec:experiments}).


\subsection{GRPO Optimization}
\label{subsec:stage2-grpo}
Building on the SFT-initialized model, we apply reinforcement learning to further optimize the policy beyond what maximum-likelihood training can achieve. As described in Section~\ref{subsec:background}, GRPO replaces the learned value network of PPO with a group-relative baseline. This design is particularly suited to our setting for three reasons:
\textit{(i)}~it avoids the memory overhead of a separate value network, which is critical when fine-tuning large multimodal models under constrained GPU resources;
\textit{(ii)}~it operates online with fresh rollouts, enabling diverse reasoning exploration;
and \textit{(iii)}~its group-relative advantage normalization naturally accommodates our composite reward function, where multiple reward components with different scales must be balanced.

The MLLM is treated as a stochastic policy $\pi_\theta$. For each input, we sample $K$ candidate outputs $\{y_{1},\dots,y_{K}\}$ and compute a scalar reward for each.
Given the structured output format defined in Section~\ref{subsec:task-formulation}, the training must balance multiple objectives: predicting the correct label, generating a faithful explanation, maintaining format compliance, and producing sufficient reasoning traces. Standard SFT provides limited control over this balance, as the cross-entropy loss treats all output tokens equally regardless of their functional role \citep{ouyang2022training}. To address this, we design a composite reward function that assigns distinct reward signals to each objective: 
\begin{align}
R(y) &= \alpha_{\text{fmt}} R_{\text{fmt}}(y) + \alpha_{\text{lbl}} R_{\text{lbl}}(y) + \alpha_{\text{exp}} R_{\text{exp}}(y) \nonumber \\
&\quad + \alpha_{\text{think}} R_{\text{think}}(y) + \alpha_{\text{met}} R_{\text{met}}(y)
\end{align}
The individual reward components are:
\begin{itemize}[noitemsep,topsep=2pt]
    \item $R_{\text{fmt}}(y)$ \textbf{(format)}: encourages structurally consistent outputs with reasoning, prediction, and explanation.
    \item $R_{\text{lbl}}(y)$ \textbf{(label correctness)}: rewards the correct class.
    \item $R_{\text{exp}}(y)$ \textbf{(explanation length)}: softly regularizes rationale length toward $\sim$100 words via 
    \[R_{\text{exp}}=\exp(-{(L-100)^2}/{2\sigma^2})\]
    where $\sigma=20$, set empirically to balance conciseness and informativeness.
    \item $R_{\text{think}}(y)$ \textbf{(thinking length)}: discourages reward-hacking behavior such as empty or uninformative CoT traces by enforcing minimum reasoning sufficiency:
    \begin{equation}
    R_{\text{think}}(y)=
    \begin{cases}
    1, & L_{\text{think}} \ge L_{\min},\\
    \exp\!\left(-\frac{(L_{\text{think}}-L_{\min})^2}{2\sigma_{\text{think}}^2}\right), & L_{\text{think}} < L_{\min},
    \end{cases}
    \end{equation}
    
    
    where $L_{\min}=150$ and $\sigma_{\text{think}}=50$ are set empirically based on observed reasoning lengths during SFT. This one-sided formulation 
    penalizes only reasoning traces shorter than the minimum threshold, thereby mitigating reward hacking \citep{firooz2025scaling} by discouraging overly short reasoning without incentivizing 
    verbosity. Without this regularization, training shows a consistent reward-hacking pattern (Appendix~\ref{app:training_dynamics}). The model learns to produce shorter or semantically empty CoT traces, compressing or omitting the \texttt{<think>} segment while still receiving high immediate reward.
    \item $R_{\text{met}}(y)$ \textbf{(semantic similarity)}: rewards METEOR similarity \citep{banerjee-lavie-2005-meteor} to the gold rationale, encouraging explanations that are both relevant and faithful to the reference annotations.
\end{itemize}

\noindent
Weights for supervised GRPO are 
set empirically on the development set that maximized macro-F1 
to $\alpha_{\text{fmt}}{=}0.35$, $\alpha_{\text{lbl}}{=}0.35$, $\alpha_{\text{exp}}{=}0.08$, $\alpha_{\text{met}}{=}0.12$, and $\alpha_{\text{think}}{=}0.10$, giving the highest priority to classification correctness and output format compliance, which are prerequisites for downstream usability.

\paragraph{Objective function.}
GRPO uses the group-average reward $\bar{R}$ as baseline:
\[
\bar{R}=\frac{1}{K}\sum_{k=1}^{K} R_k,
\qquad
A_k = \frac{R_k-\bar{R}}{\mathrm{std}(\{R_j\})+\varepsilon}
\]
Here, $A_k$ is the normalized advantage of the $k$-th completion relative to the group. Completions with above-average reward receive positive advantage and are reinforced, while below-average completions are suppressed. 
The GRPO objective maximizes the clipped surrogate with KL regularization:
\begin{align}
\mathcal{J}_{\text{G}}(\theta)
&=
\mathbb{E}\Bigg[
\sum_{k=1}^{K}\sum_{t}
\min\!\Big(
r_{k,t}A_k,\;
\mathrm{clip}(r_{k,t},1{-}\epsilon,1{+}\epsilon)A_k
\Big)
\nonumber\\
&\qquad
-\beta\,
\mathrm{KL}\!\left(
\pi_\theta(\cdot\mid h_{k,t})
\;\middle\|\;
\pi_{\text{ref}}(\cdot\mid h_{k,t})
\right)
\Bigg]
\end{align}
where $r_{k,t} = \pi_\theta(a_{k,t}\mid h_{k,t}) / \pi_{\theta_{\text{old}}}(a_{k,t}\mid h_{k,t})$ is the importance ratio, $\epsilon$ the clipping threshold that prevents excessively large policy updates, $\beta$ the KL coefficient that controls deviation from the reference policy, and $\pi_{\text{ref}}$ a fixed reference policy from the SFT checkpoint.
The clipping mechanism and KL penalty jointly ensure that each optimization step makes bounded updates to the policy, preventing training instability.

\subsection{Self-Supervised GRPO}
\label{subsec:unsup-grpo-dapo}
While the supervised GRPO pipeline described above relies on human-annotated labels, obtaining such annotations at scale is expensive and often impractical, particularly when extending to new domains or languages where labeled data is scarce. 
Recent work has explored self-supervised strategies to leverage unlabeled data during fine-tuning. MM-UPT \citep{wei2025sftsecondrlupt} introduces fully self-supervised RL using majority voting over stochastic rollouts, and V-Zero \citep{wang2026v} extends this to a co-evolutionary setting. However, these approaches have primarily targeted structured reasoning domains.
Inspired by this line of work, we extend our GRPO method to a self-supervised setting using consensus-based pseudo-supervision, and investigate whether this strategy can provide reliable supervision for meme classification, where semantic interpretation depends on subtle image--text interactions.

\paragraph{Hard-sample selection.}
Not all unlabeled samples are equally informative for RL training. Samples on which the model is already confident provide limited gradient signal, while genuinely ambiguous instances can expose decision boundaries that benefit from further optimization \citep{yang2024uncertainty}. To identify such informative instances, we run inference at multiple temperatures $\{0, 0.2, 0.4, 0.8, 1.0\}$ and retain only samples where predictions are not in full agreement, ensuring each instance exposes genuine model uncertainty.
Only samples with partial agreement (3/5 or 4/5) are retained, ensuring each training instance represents a case of genuine model uncertainty rather than random noise. 
From each dataset, we sample 2,000 such disagreement examples.

\paragraph{Consensus pseudo-reward.}
In the absence of gold labels, we derive supervision from the model's own predictions through majority voting. For each input, the model generates $K$ candidates. Let $n_c = \sum_{j=1}^{K} \mathbb{I}[\hat{c}(y_j)=c]$ and $c_{\text{maj}} = \arg\max_c n_c$. If a strict majority exists ($n_{c_{\text{maj}}} > \lfloor K/2 \rfloor$), completions predicting this class receive reward $1$; otherwise all receive $0$:
\[
R_{\text{mv}}(y_i) =
\begin{cases}
1, & n_{c_{\text{maj}}} > \lfloor K/2 \rfloor \;\text{and}\; \hat{c}(y_i) = c_{\text{maj}}, \\
0, & \text{otherwise}.
\end{cases}
\]
The strict majority requirement ensures that pseudo-labels are assigned only when the model exhibits sufficient agreement, reducing the risk of reinforcing incorrect predictions. 

The final self-supervised reward combines consensus with structural rewards:
\begin{equation}
R(y) = \alpha_{\text{fmt}} R_{\text{fmt}} + \alpha_{\text{mv}} R_{\text{mv}} + \alpha_{\text{exp}} R_{\text{exp}} + \alpha_{\text{think}} R_{\text{think}}
\end{equation}
with $\alpha_{\text{fmt}}{=}0.30$, $\alpha_{\text{mv}}{=}0.20$, $\alpha_{\text{exp}}{=}0.20$, $\alpha_{\text{think}}{=}0.30$, set empirically. 
Compared to the supervised setting, the thinking-length reward receives a higher weight to compensate for the absence of gold-label supervision, as reasoning collapse is more likely without explicit correctness feedback. 
The consensus-based pseudo-reward provides useful learning signal particularly for underrepresented categories when unlabeled data is in-domain.

\providecommand{\chg}[1]{\textcolor{blue}{#1}}

\section{Experimental Setup}
\label{sec:experiments}
This section describes the models, baselines, training configurations, and evaluation protocol used across all experiments, covering both pretrained evaluation and the multi-stage post-training pipeline.

\subsection{Models}
\label{ssec:models}
We use both open- and closed-weight MLLMs. Open-weight baselines include Llama-3.2-11B, Llama-4-Scout-17B \citep{dubey2024llama}, Qwen3-VL-8B \citep{bai2025qwen3vltechnicalreport}, Gemma-3-12B \citep{team2025gemma}, and Kimi-VL-A3B \citep{team2025kimi}. To assess whether an Arabic-centric model behaves differently on Arabic benchmark, we additionally evaluate Fanar-2-Oryx-IVU \citep{fanarteam2026fanar20arabicgenerative}. As a closed-weight model, we use GPT-4.1 \citep{openai2023gpt}. For training, we use Qwen3-VL-8B-Thinking, a reasoning-enhanced checkpoint reported in \cite{li2026qwen3} to perform strongly on benchmarks such as MMMU \citep{yue2024mmmu} and MathVista \citep{lu2024mathvista}. Experiments include both thinking and instruction-tuned variants.

\subsection{Zero-Shot and CoT Baselines}
\label{ssec:zero_shot_cot}
We evaluate all models under both zero-shot and CoT prompting. To ensure a fair comparison, we disable the internal thinking mode of thinking-enabled models in the non-CoT setting. This setup is intended to measure pretrained reasoning ability without task-specific fine-tuning. All prompts, including those used for Arabic memes, are written in English, following prior evidence that LLMs often produce more reliable and stable outputs when prompted in English \citep{kmainasi2024native}. For ArMeme, this is an evaluation design choice to improve consistency under the tested setup, not a deployment recommendation.

\subsection{Unimodal Baselines}
\label{ssec:unimodal_baselines}
To isolate the individual contribution of each modality, we fine-tune unimodal classifiers on both benchmarks. For vision, we fine-tune BEiT~\citep{bao2022beit}, ConvNeXt~\citep{liu2022convnet}, DINOv2~\citep{oquab2024dinov2}, ResNet-101~\citep{he2016resnet}, Swin Transformer~\citep{liu2021swin}, and ViT~\citep{dosovitskiy2021vit}. For text, we fine-tune mBERT~\citep{devlin2019bert}, QARiB~\citep{abdelali2021qarib}, AraBERTv2~\citep{antoun2020arabert}, DistilBERT~\citep{sanh2019distilbert}, and XLM-RoBERTa~\citep{conneau2020xlmr}. All models are trained under a unified setup to ensure fair comparison.
 
\subsection{Multimodal Sequence Classification Baselines}
\label{ssec:seq_cls_baselines}
To contextualize the performance of our generative RL-based approach, we fine-tune Gemma-3-12B-IT~\citep{kamath2025gemma3} and Qwen3-VL-8B-Instruct~\citep{yang2025qwen3} on both training splits of both benchmarks, and additionally Qwen3-VL-8B-Thinking~\citep{yang2025qwen3} on ArMeme, using standard cross-entropy classification heads. Unlike our generative models, these baselines are trained solely on the classification objective and do not generate explanations.

\subsection{Fine-Tuning Configurations}
\label{ssec:ft_configs}
This section describes the configurations for the three post-training stages introduced in Section~\ref{sec:methodology} and also presented in Figure \ref{fig:overview}: \textit{(i)} SFT warm-up, \textit{(ii)} supervised GRPO, and \textit{(iii)} self-supervised GRPO. 

\paragraph{SFT warm-up}
The three SFT variants are described in Section~\ref{subsec:sft-warmup}. In variants without CoT, empty \texttt{<think></think>} tags are included to preserve the thinking format while excluding them from loss computation.

\paragraph{Supervised GRPO}
We apply supervised GRPO from multiple initialization checkpoints to analyze the effect of SFT warm-up. 
We compare GRPO initialized directly from the pretrained backbone, referred as \texttt{cold-start}, with GRPO initialized from task-specific SFT checkpoints, thereby measuring the contribution of supervised warm-up before reinforcement learning. We also evaluate the effect of the thinking-length reward $R_{\text{think}}$.

\paragraph{Self-supervised GRPO.}
Self-supervised GRPO is initialized from the best supervised checkpoint and trained on 2,000 disagreement-based unlabeled samples per dataset (Section~\ref{subsec:unsup-grpo-dapo}).

\subsection{Training Setup}
\label{ssec:training_setup}
We describe the key training hyperparameters below; full configurations for all stages, including unimodal baselines, are provided in Appendix~\ref{app:training_configs}. 
All multimodal models are trained with partial parameter fine-tuning, where only the language model is updated while visual components remain frozen.
Optimization was done using DeepSpeed ZeRO-3 in bfloat16 on 4 NVIDIA H200 GPUs, with gradient checkpointing and the vision encoder frozen. SFT uses a per-device batch size of 4, learning rate $1{\times}10^{-5}$, cosine scheduler with warm-up ratio 0.05, and AdamW with weight decay 0.1 for 5 epochs. GRPO uses learning rate $1{\times}10^{-6}$, batch size 4, KL coefficient 0.05, clip range 0.2, temperature 1.0, top-$p$ 0.85, and $K{=}16$ candidates per input with max 4096 tokens for 5 epochs. Self-supervised GRPO uses learning rate $2{\times}10^{-7}$, gradient accumulation of 4, and trains for 4 epochs. Unless otherwise noted, the reported results are from a single random seed (42).

\subsection{Evaluation Metrics}
\label{ssec:eval_metrics}
We follow the evaluation protocol of \cite{kmainasi2025memeintel}. For the classification task, we report Accuracy (Acc), Weighted F1 (W-F1), and Macro F1 (M-F1). For explanation quality, we report BERTScore (BS) \citep{zhang2020bertscoreevaluatingtextgeneration} and METEOR (MET) \citep{banerjee-lavie-2005-meteor}. We use these as reference-based measures of explanation similarity, rather than as direct indicators of reasoning faithfulness or logical validity.

We use \textit{macro-F1 as the primary comparison metric} because it is more robust to class imbalance, while also reporting accuracy for comparability with prior work.

\providecommand{\chg}[1]{\textcolor{blue}{#1}}

\section{Results and Analysis}
\label{sec:results_analysis}
This section presents our experimental results. We begin with a comparison against prior work, followed by unimodal and zero-shot baselines, a detailed analysis of RL post-training and self-supervised GRPO, cross-dataset analysis, statistical significance tests, and a summary of findings organized by research question.

\subsection{Comparison with State-of-the-Art}
\label{ssec:sota_comparison}

Tables~\ref{tab:sota_fhm} and~\ref{tab:sota_armeme} compare our approach with published baselines.\footnote{All baseline results are reported from the original papers and are not re-implemented.}
On FHM (Table~\ref{tab:sota_fhm}), our approach achieves 0.80 macro-F1, outperforming the multimodal sequence-classification baselines Gemma-3-12B-IT and Qwen3-VL-8B-Instruct, which obtain lower macro-F1 and do not generate explanations. This indicates that our reasoning-based approach improves balanced classification performance while also producing interpretable rationales.

On ArMeme (Table~\ref{tab:sota_armeme}), self-supervised GRPO achieves 0.612 macro-F1, improving over \cite{kmainasi2025memeintel} by +7.6 points and over the original ArMeme benchmark of \cite{alam-etal-2024-armeme} by +6.1 points. It also outperforms the strongest sequence-classification baseline, Qwen3-VL-8B-Instruct, in macro-F1 (0.612 vs.\ 0.594), while additionally generating explanations. These results highlight the advantage of reasoning-based optimization for balanced per-class performance, particularly in the more challenging four-class ArMeme setting.

\begin{table}[t]
\centering
\small
\setlength{\tabcolsep}{2pt}
\caption{Comparison with SOTA on \textbf{Hateful Memes}. Seq.\ cls.\ baselines use classification heads without explanations. Sup.: Supervised GRPO.}
\label{tab:sota_fhm}
\begin{tabular}{lccccc}
\toprule
\textbf{Model} & \textbf{Acc} & \textbf{W-F1} & \textbf{M-F1} & \textbf{BS} & \textbf{MET} \\
\midrule
\multicolumn{6}{c}{\textit{Prior work}} \\
\midrule
\cite{kiela2020hateful}          & 69.5 & --   & --   & --   & --   \\
\cite{cao-etal-2022-prompting}   & 73.0 & --   & --   & --   & --   \\
\cite{wu2024multimodal}          & 76.4 & --   & --   & --   & --   \\
\cite{yang2024uncertainty}       & 77.2 & --   & --   & --   & --   \\
\cite{burbi2023mapping}          & 77.7 & --   & --   & --   & --   \\
\cite{mei-etal-2024-improving}          & 78.8 & --   & --   & --   & --   \\
\cite{kmainasi2025memeintel}     & 79.9 & 0.80 & 0.79 & 0.78 & 0.49 \\
\midrule
\multicolumn{6}{c}{\textit{Seq.\ cls.\ baselines}} \\
\midrule
Gemma-3-12B-IT        & 78.0 & 0.77 & 0.74 & -- & -- \\
Qwen3-VL-8B-Instruct  & 78.0 & 0.78 & 0.77 & -- & -- \\
\midrule
\multicolumn{6}{c}{\textit{Proposed}} \\
\midrule
Proposed (Sup.)           & \textbf{82.0} & \textbf{0.82} & \textbf{0.80} & \textbf{0.78} & \textbf{0.52} \\
\quad + Self-Sup.                & 81.8 & 0.81 & 0.79 & 0.77 & 0.51 \\

\bottomrule
\end{tabular}
\end{table}

\begin{table}[t]
\centering
\small
\setlength{\tabcolsep}{2pt}
\caption{Comparison with SOTA on \textbf{ArMeme}. Seq.\ cls.\ baselines use classification heads without explanations. Sup.: Supervised GRPO. Self-Sup.: Self-supervised GRPO.}
\label{tab:sota_armeme}
\begin{tabular}{lccccc}
\toprule
\textbf{Model} & \textbf{Acc} & \textbf{W-F1} & \textbf{M-F1} & \textbf{BS} & \textbf{MET} \\
\midrule
\multicolumn{6}{c}{\textit{Prior work}} \\
\midrule
\cite{alam-etal-2024-armeme}     & 69.7 & 0.69 & .551 & -- & -- \\
\cite{kmainasi2025memeintel}     & 72.1 & .699 & .536 & .70 & .35 \\
\midrule
\multicolumn{6}{c}{\textit{Seq.\ cls.\ baselines}} \\
\midrule
Gemma-3-12B-IT        & 74.8 & .724 & .597 & -- & -- \\
Qwen3-VL-8B-Instruct  & \textbf{76.6} & \textbf{.753} & .594 & -- & -- \\
\midrule
\multicolumn{6}{c}{\textit{Proposed}} \\
\midrule
Proposed (Sup.)       & 72.6 & .711 & .597 & .740 & .441 \\
\quad + Self-Sup.  & 72.8 & .715 & \textbf{.612} & \textbf{.741} & \textbf{.442} \\
\bottomrule
\end{tabular}
\end{table}

\subsection{Unimodal Baselines}
\label{ssec:unimodal_results}


As unimodal baselines, we fine-tune image-only and text-only classifiers on both benchmarks
(Table~\ref{tab:unimodal_combined}). Across both datasets, text-only classifiers consistently
outperform image-only classifiers in macro-F1. The gap is especially large on ArMeme, where
mean macro-F1 increases from 0.267 for image-only models to 0.482 for text-only models,
suggesting that textual cues provide the strongest signal for propaganda detection.

However, unimodal models still lag behind our multimodal reasoning approach. On FHM, the
best unimodal baseline reaches 0.53 macro-F1, compared with 0.80 for our approach. On
ArMeme, the gap is 0.509 vs. 0.612 macro-F1. These results show that cross-modal reasoning captures complementary signals that unimodal models often miss. This is especially important for memes, where harmful content often emerges from the interaction between the image and text.

\begin{table}[t]
\centering
\small
\setlength{\tabcolsep}{2pt}
\caption{\textbf{Unimodal fine-tuned baselines on Hateful Memes and ArMeme.} Best per column in \textbf{bold}; ``--'' indicates the model was not evaluated on that dataset.}
\label{tab:unimodal_combined}
\begin{tabular}{lccc|ccc}
\toprule
& \multicolumn{3}{c}{\textbf{FHM}} & \multicolumn{3}{|c}{\textbf{ArMeme}} \\
\cmidrule(lr){2-4}\cmidrule(lr){5-7}
\textbf{Model} & \textbf{Acc} & \textbf{W-F1} & \textbf{M-F1} & \textbf{Acc} & \textbf{W-F1} & \textbf{M-F1} \\
\midrule
\multicolumn{7}{c}{\textit{Image-only}} \\
\midrule
BEiT-base & 61.0 & .560 & .500 & 65.1 & .578 & .289 \\
ConvNeXt-large & \textbf{63.0} & .510 & .420 & 65.5 & .527 & .203 \\
DINOv2-large & 58.0 & .560 & \textbf{.510} & 65.8 & \textbf{.618} & \textbf{.402} \\
ResNet-101 & \textbf{63.0} & .480 & .380 & 65.8 & .522 & .198 \\
Swin-large & \textbf{63.0} & \textbf{.570} & \textbf{.510} & \textbf{67.2} & .585 & .267 \\
ViT-base & 61.0 & .560 & .500 & 64.0 & .556 & .243 \\
\midrule
\textbf{Average} & 61.5 & .540 & .470 & 65.6 & .564 & .267 \\
\midrule
\multicolumn{7}{c}{\textit{Text-only}} \\
\midrule
AraBERTv2 & -- & -- & -- & 67.9 & .668 & .503 \\
DistilBERT & \textbf{64.0} & \textbf{.590} & \textbf{.530} & -- & -- & -- \\
mBERT & 64.0 & .550 & .480 & 71.3 & .676 & .446 \\
Qarib & 63.0 & .570 & .520 & 70.5 & .670 & .469 \\
XLM-RoBERTa & 64.0 & .570 & .510 & \textbf{71.5} & \textbf{.691} & \textbf{.509} \\
\midrule
\textbf{Average} & 63.8 & .570 & .510 & 70.3 & .676 & .482 \\
\bottomrule
\end{tabular}
\end{table}

\subsection{Zero-Shot and CoT}
\label{sec:zero_shot_eval}

Table~\ref{tab:pretrained_mllms} reports the zero-shot and CoT results across all models on both benchmarks. For clarity, we use abbreviated model names: \textbf{Llama} (Llama-3.2-11B)~\citep{dubey2024llama}, \textbf{L4-Scout} (Llama-4-Scout-17B),\footnote{\url{https://ai.meta.com/blog/llama-4-multimodal-intelligence/}} \textbf{Qwen-I/T} (Qwen3-VL-8B Instruct/Thinking), \textbf{Gemma} (Gemma-3-12B), and \textbf{Kimi-I/T} (Kimi-VL-A3B Instruct/Thinking). 

On FHM, no single model dominates: GPT-4.1~\citep{openai2023gpt4} with CoT the best macro-F1, and \textbf{L4-Scout} obtains best BS score.
%
On ArMeme, all models exhibit substantially lower performance compared to binary FHM dataset, reflecting the added difficulty of four-class Arabic propaganda classification. 
Overall, most models underperform without CoT on both tasks. This trend is clearer on ArMeme. For example, Llama without CoT predicts the \textit{not-meme} class for roughly 95\% of test instances and achieves only 7.0\% accuracy.



\begin{table*}[t]
\centering
\small
\setlength{\tabcolsep}{3pt}
\caption{\textbf{Performance of pretrained MLLMs on both benchmarks.}
Llama = Llama-3.2-11B; L4-Scout = Llama-4-Scout-17B; Qwen-I/T = Qwen3-VL-8B (Instruct/Thinking); Gemma = Gemma-3-12B; Kimi-I/T = Kimi-VL-A3B (Instruct/Thinking). Acc: Accuracy; BS: BERTScore; Met: METEOR.}
\label{tab:pretrained_mllms}
\begin{tabular}{l ccccc|ccccc}
\toprule
& \multicolumn{5}{c|}{\textbf{FHM}} & \multicolumn{5}{c}{\textbf{ArMeme}} \\
\cmidrule(lr){2-6} \cmidrule(lr){7-11}
\textbf{Model} & \textbf{Acc} & \textbf{W-F1} & \textbf{M-F1} & \textbf{BS} & \textbf{Met} & \textbf{Acc} & \textbf{W-F1} & \textbf{M-F1} & \textbf{BS} & \textbf{Met} \\
\midrule
Llama +CoT    & 60.0 & .606 & .592 & .600 & .165   & 35.3 & .401 & .216 & .563 & .131 \\
Llama         & 64.2 & .599 & .547 & .660 & .234   &  7.0 & .048 & .061 & .591 & .157 \\
\midrule
Qwen-I +CoT   & 65.3 & .658 & .652 & .646 & \textbf{.263}  & 64.6 & .626 & .388 & .627 & \textbf{.210} \\
Qwen-I        & 63.0 & .630 & .630 & .662 & .216   & 30.2 & .373 & .259 & .616 & .199 \\
\midrule
Qwen-T +CoT   & 70.2 & .706 & .694 & .581 & .202   & 63.1 & .561 & .281 & .598 & .173 \\
Qwen-T        & 65.6 & .661 & .652 & .659 & .215   & 22.1 & .108 & .168 & .601 & .207 \\
\midrule
Gemma +CoT    & 65.5 & .659 & .655 & .621 & .135   & \textbf{67.5} & \textbf{.625} & .345 & .597 & .103 \\
Gemma         & 65.2 & .654 & .652 & .664 & .215   & 28.5 & .197 & .155 & .627 & .213 \\
\midrule
L4-Scout +CoT & 75.4 & .687 & .545 & .596 & .111   & 67.9 & .582 & .284 & .589 & .093 \\
L4-Scout      & \textbf{77.0} & .710 & .580 & .653 & .252   & 54.1 & .569 & .340 & \textbf{.649} & \textbf{.276} \\
\midrule
GPT-4.1 +CoT  & 74.1 & \textbf{.739} & \textbf{.719} & .470 & .072   & 68.7 & .599 & .290 & .577 & .083 \\
GPT-4.1       & 71.3 & .715 & .700 & .595 & .159   & \textbf{69.7} & \textbf{.619} & \textbf{.350} & .637 & .180 \\
\midrule
Kimi-I +CoT   & 67.7 & .667 & .636 & .621 & .186   & 39.9 & .437 & .235 & .602 & .134 \\
Kimi-I        & 67.0 & .625 & .573 & \textbf{.678} & .257   & 21.8 & .263 & .176 & .618 & .200 \\
\midrule
Kimi-T +CoT   & 70.5 & .704 & .685 & .614 & .155   & 57.2 & .562 & .310 & .586 & .115 \\
Kimi-T        & 67.5 & .668 & .639 & .660 & .222   & 28.3 & .336 & .217 & .611 & .223 \\
\midrule
Fanar-2-Oryx-IVU +CoT & 68.0 & .680 & .660 & .600 & .120 & 64.4 & .593 & .330 & .582 & .091 \\
Fanar-2-Oryx-IVU      & 69.0 & .690 & .670 & .650 & .200 & 57.3 & .504 & .240 & .614 & .161 \\
\bottomrule
\end{tabular}
\end{table*}

\subsubsection{Impact of CoT}

Averaging across all nine models, CoT prompting improves mean macro-F1 by $+0.022$ on FHM and $+0.079$ on ArMeme, while consistently degrading explanation quality (mean BS drops of $-5.91$ and $-2.7$ points, respectively).

On FHM, CoT exhibits a \emph{systematic and asymmetric effect}: it improves classification for most models (up to +0.063 Macro-F1 points for Kimi-I), primarily by reducing false negatives, but \emph{consistently degrades explanation quality} as measured by BS (drops of 1.6 to 12.5 points), suggesting that CoT shifts explanations away from the concise annotation style used in the references.

On ArMeme, CoT produces \emph{dramatically larger} Macro-F1 gains (up to +0.19 for Gemma), demonstrating that explicit reasoning is essential when models face a multilingual, multi-class task far from their pretraining distribution. The sole exception is GPT-4.1, which performs slightly better without CoT, suggesting its internal representations already encode sufficient task knowledge for Arabic propaganda detection. Across both benchmarks, CoT benefits vary substantially across models rather than being confined to a specific model family, while explanation quality consistently degrades, highlighting a trade-off between classification performance and explanation calibration.


\paragraph{Model selection.}
These observations demonstrate the need for further investigation into how to effectively exploit open-weight thinking-based MLLMs that are trainable under limited computational resource. Based on overall performance across both benchmarks and ease of implementation, we select the Qwen-T
model as the base MLLM for our subsequent experiments.

\subsection{Analysis of RL Post-Training}
\label{sec:rl_analysis}

Tables~\ref{tab:sft_grpo_fhm} and~\ref{tab:sft_grpo_armeme} report the effect of SFT warm-up and GRPO on both benchmarks. We organize the analysis around four key findings.

\begin{table}[t]
\centering
\small
\setlength{\tabcolsep}{2pt}
\caption{\textbf{SFT and GRPO on FHM.} Cls = class labels; FG = fine-grained; Exp = explanation; CoTD = CoT distillation; $R_{\text{think}}$ = thinking-length reward. $^{\dagger}$/$^{\ddagger}$ denote Holm-corrected paired-bootstrap improvements in macro-F1 over the SFT warm-up the row was initialised from, at $\alpha\!=\!0.05$/$0.01$ (one-sided); see Section~\ref{sec:sig_test}.}
\label{tab:sft_grpo_fhm}
\begin{tabular}{lccccc}
\toprule
\textbf{Model} & \textbf{Acc} & \textbf{W-F1} & \textbf{M-F1} & \textbf{BS} & \textbf{Met} \\
\midrule
\multicolumn{6}{c}{\textit{SFT warm-up}} \\
\midrule
SFT (Cls+Exp)              & 77.0 & 0.77 & 0.75 & 0.77 & 0.48 \\
SFT (Cls+FG+Exp)           & 78.1 & 0.78 & 0.77 & 0.77 & 0.48 \\
SFT (Cls+FG+Exp, CoTD)     & 79.2 & 0.79 & 0.78 & 0.78 & 0.50 \\
\midrule
\multicolumn{6}{c}{\textit{RL / GRPO}} \\
\midrule
GRPO (Cold Start)                            & 76.8 & 0.77 & 0.75 & 0.73 & 0.47 \\
SFT-Cls+Exp $\to$ GRPO$^{\ddagger}$          & 80.4 & 0.80 & 0.78 & 0.76 & 0.50 \\
SFT-Cls+FG+Exp $\to$ GRPO$^{\dagger}$        & 81.1 & 0.81 & 0.79 & 0.77 & 0.52 \\
SFT-CoTD $\to$ GRPO                          & 81.2 & 0.81 & 0.79 & 0.78 & 0.52 \\
\midrule
\multicolumn{6}{c}{\textit{RL / GRPO + $R_{\text{think}}$}} \\
\midrule
SFT-CoTD $\to$ GRPO + $R_{\text{think}}$\textsuperscript{$\dagger$}
                            & {82.0} & {0.82} & {0.80} & {0.78} & {0.52} \\
\bottomrule
\end{tabular}
\end{table}

\begin{table}[t]
\centering
\small
\setlength{\tabcolsep}{2pt}
\caption{\textbf{SFT and GRPO on ArMeme.} Same notation as Table~\ref{tab:sft_grpo_fhm}; $^{\dagger}$/$^{\ddagger}$ denote Holm-corrected paired-bootstrap macro-F1 gains over the corresponding SFT warm-up (Section~\ref{sec:sig_test}).}
\label{tab:sft_grpo_armeme}
\begin{tabular}{lccccc}
\toprule
\textbf{Model} & \textbf{Acc} & \textbf{W-F1} & \textbf{M-F1} & \textbf{BS} & \textbf{Met} \\
\midrule
\multicolumn{6}{c}{\textit{SFT warm-up}} \\
\midrule
SFT (Cls+Exp)              & 64.2 & 0.63 & 0.43 & 0.61 & 0.35 \\
SFT (Cls+FG+Exp)           & 69.3 & 0.67 & 0.51 & 0.73 & 0.43 \\
SFT (Cls+FG+Exp, CoTD)     & 70.3 & 0.69 & 0.54 & 0.73 & 0.43 \\
\midrule
\multicolumn{6}{c}{\textit{RL / GRPO}} \\
\midrule
GRPO (Cold Start)                              & 70.9 & .651 & .332 & .711 & .421 \\
SFT-Cls+FG+Exp $\to$ GRPO$^{\dagger}$          & 72.4 & .703 & .557 & .741 & .439 \\
SFT-Cls+Exp $\to$ GRPO$^{\ddagger}$            & 72.8 & .706 & .579 & .742 & .443 \\
SFT-CoTD $\to$ GRPO$^{\dagger}$                & 72.9 & .706 & .577 & .741 & .441 \\
\midrule
\multicolumn{6}{c}{\textit{RL / GRPO + $R_{\text{think}}$}} \\
\midrule
SFT-CoTD $\to$ GRPO + $R_{\text{think}}$\textsuperscript{$\dagger$}
                                     & 72.6 & .711 & .597 & .740 & .441 \\
\bottomrule
\end{tabular}
\end{table}


On FHM, SFT with fine-grained supervision (Cls + FG + Exp) improves macro-F1 from 0.75 to 0.77 over the variant without fine-grained labels. The gain is substantially larger on ArMeme, where macro-F1 increases by +8.0\%. This shows that incorporating fine-grained labels improves decision consistency even without explicit reasoning supervision. The effect is particularly clear on ArMeme, where our multi-LLM annotation pipeline provides effective fine-grained supervision for a dataset that previously lacked such annotations.


\paragraph{CoT distillation improves explanation quality.}
Adding distilled CoT supervision during SFT yields further gains (Tables~\ref{tab:sft_grpo_fhm},~\ref{tab:sft_grpo_armeme}). SFT (Cls+FG+Exp, CoTD) achieves the strongest warm-up performance on FHM, reaching 0.78 macro-F1, a +1 point gain over SFT (Cls+FG+Exp), with a 0.01 increase in BS. On ArMeme, it improves macro-F1 from 0.51 to 0.54, while BS remains unchanged. Thus, the effect of CoTD on explanation quality is more visible on FHM during SFT. For ArMeme, improvements in BS emerge only after GRPO. Overall, explicit reasoning supervision improves classification performance on both datasets and provides a stronger initialization for GRPO.

\paragraph{SFT warm-up is critical for GRPO.}
Cold-start GRPO achieves only 0.75 and 0.33 macro-F1 on FHM and ArMeme, respectively, showing that GRPO provides limited benefits without proper warm-up. This variant also produces weaker explanations, with a BS of 0.73 on FHM, further highlighting the importance of SFT initialization. In contrast, all SFT-initialized GRPO models substantially outperform both SFT-only and cold-start variants. For example, initializing GRPO from SFT-Cls+FG+Exp improves macro-F1 from 0.77 to 0.79 on FHM and from 0.51 to 0.58 on ArMeme. These results suggest that once the model is aligned with the target task through SFT, GRPO can more effectively refine label correctness and rationale quality.


\paragraph{Thinking-length regularization.}
Adding $R_{\text{think}}$ to the CoTD-initialized variant improves macro-F1 on both benchmarks
(Tables~\ref{tab:sft_grpo_fhm},~\ref{tab:sft_grpo_armeme}). On FHM, macro-F1 increases from
0.79 to 0.80, yielding the strongest supervised result. On ArMeme, the gain is larger, with
macro-F1 improving from 0.57 to 0.59 (+2.0\%). This suggests that $R_{\text{think}}$ mainly
improves balanced per-class performance, especially on the harder four-class ArMeme task.
This is consistent with the training-dynamics analysis below: $R_{\text{think}}$ stabilizes
reasoning length and mitigates length-shortening bias in rationale generation.

\paragraph{GRPO training dynamics.}
An analysis of reward trajectories and completion lengths across initialization regimes (see Figure~\ref{fig:reward_dynamics} in Appendix~\ref{app:training_dynamics}) reveals a consistent \textit{reward hacking} phenomenon, particularly on ArMeme. The mean completion length decreases rapidly while the reward continues to increase, indicating that the policy exploits a shortcut by producing shorter generations rather than improving reasoning quality. Initialization with CoTD partially mitigates this behavior by introducing a structural prior over reasoning generation, while the explicit $R_{\text{think}}$ reward stabilizes training dynamics across both datasets. These observations further support the importance of SFT warm-up and thinking-length regularization reported above.


\subsection{Self-Supervised GRPO}
\label{ssec:self_sup_results}
We evaluate self-supervised GRPO (Section~\ref{subsec:unsup-grpo-dapo}) by further training the best supervised GRPO + $R_{\text{think}}$ checkpoint on 2,000 unlabeled memes per dataset. The memes are selected using disagreement-based sampling under multi-temperature inference. We retain only samples with partial agreement (3/5 or 4/5), so each training instance contains measurable prediction uncertainty. Table~\ref{tab:self_sup_grpo} reports the effect of this self-supervised stage.

\begin{table}[t]
\centering
\small
\setlength{\tabcolsep}{3pt}
\caption{\textbf{Effect of self-supervised GRPO} The supervised baseline is the best GRPO + $R_{\text{think}}$ checkpoint. Sup.: Supervised. $^{\dagger}$ denotes a Holm-corrected paired-bootstrap improvement of the self-supervised row over the supervised baseline at $\alpha\!=\!0.05$ (one-sided); see Section~\ref{sec:sig_test}.}
\label{tab:self_sup_grpo}
\begin{tabular}{lccccc}
\toprule
\textbf{Model} & \textbf{Acc} & \textbf{W-F1} & \textbf{M-F1} & \textbf{BS} & \textbf{Met} \\
\midrule
\multicolumn{6}{c}{\textit{FHM}} \\
\midrule
Sup.\ GRPO + $R_{\text{think}}$        & \textbf{82.0} & \textbf{0.82} & \textbf{0.80} & \textbf{0.78} & \textbf{0.52} \\
\quad + Self-Sup.\ GRPO                & 81.8 & 0.81 & 0.79 & 0.77 & 0.51 \\
\midrule
\multicolumn{6}{c}{\textit{ArMeme}} \\
\midrule
Sup.\ GRPO + $R_{\text{think}}$            & 72.6 & .711 & .597 & .740 & .441 \\
\quad + Self-Sup.\ GRPO$^{\dagger}$        & \textbf{72.8} & \textbf{.715} & \textbf{.612} & \textbf{.741} & \textbf{.442} \\
\bottomrule
\end{tabular}
\end{table}


\paragraph{ArMeme.}
Self-supervised GRPO improves macro-F1 by 1.5 points, from 0.597 to 0.612, and yields a small gain in METEOR, from .441 to .442 (Table~\ref{tab:self_sup_grpo}). A per-class recall analysis shows that the gains are concentrated in minority classes. Propaganda improves from 40.0\% to 41.8\%, not-meme from 54.4\% to 57.9\%, and other from 50.0\% to 51.8\%. 
This comes at a small cost to the majority class, not-propaganda, which decreases by 0.8 points. The prediction distribution shifts slightly toward minority classes and moves closer to the true label distribution. These results suggest that the consensus-based pseudo-reward provides a useful learning signal for underrepresented categories when unlabeled data is in-domain (see Figure~\ref{fig:qual_ex2} for an example).

\paragraph{FHM.}
On FHM, self-supervised GRPO reduces macro-F1 by 1.0 point (Table~\ref{tab:self_sup_grpo}). Per-class analysis shows an asymmetric effect. Not-hateful recall improves from 87.5\% to 92.2\%, while hateful recall drops from 72.9\% to 64.4\%. The model develops a bias toward the majority class, with 80.5\% of the case model predicted the meme as not-hateful. The different outcomes on the two benchmarks are not contradictory. On the four-class ArMeme task, majority consensus is harder to reach, so the pseudo-reward fires more often on borderline cases, including minority-class examples. This provides gradient signal for underrepresented categories. On the binary FHM task, small policy biases can more easily produce unanimous agreement on the majority class. As a result, the pseudo-reward becomes self-reinforcing for ``not-hateful'' and amplifies, rather than corrects, the existing class skew.

\paragraph{Analysis of dataset-dependent behavior.}
We attribute the divergent outcomes to three factors.

\textit{(i)}~\textit{Distribution alignment}: The ArMeme unlabeled set is collected from the same Arabic social media sources as the labeled set, ensuring distributional consistency, whereas the FHM unlabeled set is curated from different English meme datasets, introducing a potential distribution shift relative to the FHM benchmark.

\textit{(ii)}~\textit{Majority-vote bias amplification}: In binary classification, even small prediction biases can produce near-unanimous agreement among sampled outputs, leading to uniform rewards and weak gradient signals. In contrast, the four-class setting makes majority consensus harder to achieve, preserving reward variation and gradient diversity.

\textit{(iii)}~\textit{Task complexity}: Binary tasks rely on a single decision boundary, which can be quickly shifted by biased pseudo-supervision, whereas multi-class tasks involve multiple decision boundaries and provide richer gradient signals.

\subsection{Cross-Dataset Analysis}
\begin{table}[t]
\centering
\small
\setlength{\tabcolsep}{3pt}
\caption{\textbf{Cross-dataset comparison.} FHM = Hateful Memes (binary); ArMeme = Arabic propaganda (four-class).}
\label{tab:cross_dataset}
\begin{tabular}{lccc|ccc}
\toprule
& \multicolumn{3}{c|}{\textbf{FHM}} & \multicolumn{3}{c}{\textbf{ArMeme}} \\
\cmidrule(lr){2-4} \cmidrule(lr){5-7}
\textbf{Setting} & \textbf{Acc} & \textbf{M-F1} & \textbf{BS} & \textbf{Acc} & \textbf{M-F1} & \textbf{BS} \\
\midrule
Best unimodal (image)    & 63.0 & .510  & --   & 67.2 & .267 & --   \\
Best unimodal (text)     & 64.0 & .530  & --   & 71.5 & .509 & --   \\
Best zero-shot           & 77.0 & .719 & .678 & 69.7 & .350 & .649 \\
Best seq.\ cls.          & 78.0 & .770  & --   & \textbf{76.6} & .594 & --   \\
Best SFT                 & 79.2 & .780  & .780  & 70.3 & .54  & .73  \\
Best GRPO                & 81.2 & .790  & .780  & 72.9 & .577 & .741 \\
GRPO+$R_{\text{think}}$  & \textbf{82.0} & \textbf{.800} & \textbf{.780} & 72.6 & .597 & .740 \\
+ Self-Sup.              & 81.8 & .790  & .770  & 72.8 & \textbf{.612} & \textbf{.741} \\
\midrule
Prior SOTA               & 79.9 & .790  & .780  & 72.1 & .536 & .70  \\
\bottomrule
\end{tabular}
\end{table}
Table~\ref{tab:cross_dataset} provides a side-by-side comparison of the best results under each experimental setting.
Several cross-dataset patterns emerge from this analysis.
\paragraph{Task difficulty.}
ArMeme is consistently more challenging than FHM across all experimental settings. In the zero-shot setting, macro-F1 drops from 0.719 on FHM to 0.350 on ArMeme. This gap reflects the combined effect of four-class classification (vs.\ binary classification), Arabic-language difficulties, and the inherent subtlety of propaganda detection compared with hatefulness detection.



\paragraph{Effectiveness of RL post-training.}
On both benchmarks, performance improves from SFT warm-up to GRPO and GRPO+$R_{\text{think}}$ (Tables~\ref{tab:sft_grpo_fhm},~\ref{tab:sft_grpo_armeme}). Compared with the best SFT checkpoint, SFT-CoTD, the best supervised GRPO+$R_{\text{think}}$ checkpoint improves macro-F1 from .78 to .80 on FHM and from .54 to .597 on ArMeme. The gain is therefore larger on ArMeme, the more imbalanced four-class task. This suggests that RL-based post-training is useful for improving balanced per-class performance in our setting.


\paragraph{RL without supervised warm-up.}
When GRPO is applied directly to the base model, without an initial SFT stage, performance on ArMeme collapses to a degenerate macro-F1 of 0.332, while remaining reasonable on FHM (0.75), see Tables~\ref{tab:sft_grpo_fhm} and ~\ref{tab:sft_grpo_armeme}. This shows that starting RL from an unadapted base model is much more fragile as task complexity increases, making SFT initialization particularly important for challenging multilingual and multi-class settings.

\paragraph{Generative reasoning vs.\ classification.}
On ArMeme, the best seq.\ cls.\ baseline has lower macro-F1 than our generative model (0.594 vs.\ 0.612) and produces no explanations. On FHM, our approach also outperforms seq.\ cls.\ baselines in macro-F1. This contrast suggests that generative reasoning models can provide more balanced per-class performance while also producing interpretable outputs.

\subsection{Statistical Significance}
\label{sec:sig_test}

We assess statistical significance using paired bootstrap resampling over test examples \citep{dror2018hitchhiker} with $B=10{,}000$ samples. Since all models are evaluated on the same test sets, each bootstrap sample preserves the paired predictions by example index and recomputes macro-F1 for both systems. We use a one-sided alternative hypothesis, testing whether the post-training model improves over its comparison checkpoint. Each post-training row in Tables~\ref{tab:sft_grpo_fhm}, \ref{tab:sft_grpo_armeme}, and \ref{tab:self_sup_grpo} is compared against the checkpoint indicated by the row label, such as GRPO vs. its SFT warm-up, or self-supervised GRPO vs. the preceding supervised GRPO. Cold-start GRPO is excluded. We control the family-wise error rate within each table using Holm--Bonferroni correction \citep{holm1979simple}.

For \textbf{ArMeme} and \textbf{FHM}, supervised GRPO improvements are statistically significant at $p \leq 0.05$ after correction, except \texttt{SFT-CoTD\,$\to$\,GRPO} on FHM (Holm-adjusted $p=0.12$). In Table~\ref{tab:self_sup_grpo}, the FHM self-supervised result is not significant ($p=0.94$), while the ArMeme result is marginally significant ($p=0.047$). Overall, supervised RL provides stronger evidence of improvement over SFT initialization, whereas the self-supervised stage shows mixed evidence.

\subsection{Summary of Findings by Research Question}
\label{ssec:rq_discussion}

We revisit the research questions posed in the introduction and summarize our findings.

\paragraph{RQ1. Fine-grained supervision.}
On FHM (Table~\ref{tab:sft_grpo_fhm}), fine-grained labels provide consistent gains. SFT-Cls+FG+Exp $\to$ GRPO reaches .79 macro-F1, compared with .78 for SFT-Cls+Exp $\to$ GRPO without fine-grained supervision. On ArMeme (Table~\ref{tab:sft_grpo_armeme}), the effect is largest during SFT. SFT (Cls+FG+Exp) improves over SFT (Cls+Exp) from .43 to .51 macro-F1, an +8.0 point improvement.


\paragraph{RQ2: Synthetic CoT supervision.} CoTD provides consistent gains during SFT warm-up on both benchmarks: +1 point macro-F1 and +2\% METEOR on FHM; +3 points macro-F1 and comparable METEOR score on ArMeme. When used as GRPO initialization, CoTD-based checkpoints achieve the highest BERTScore (0.78 on FHM), suggesting that explicit reasoning supervision can improve reference-based explanation fidelity and provides a stronger starting point for RL.

\paragraph{RQ3. Multi-LLM synthetic annotation.}
Our annotation pipeline, using GPT-4.1 and Llama-4-Scout as annotators and Gemini-3-Pro as consolidator, provides effective propaganda supervision on ArMeme, where no prior fine-grained annotations were available. At the SFT stage, it improves macro-F1 from .43 to .51, an +8.0 point improvement. This suggests that multi-LLM pipelines can provide scalable proxy fine-grained supervision in our setting.

\paragraph{RQ4. Supervised GRPO post-training.}
GRPO consistently improves over SFT baselines on both datasets (Tables~\ref{tab:sft_grpo_fhm},~\ref{tab:sft_grpo_armeme}). On FHM, the best SFT checkpoint, CoTD, reaches .78 macro-F1, while the best supervised GRPO+$R_{\text{think}}$ checkpoint reaches .80. On ArMeme, macro-F1 improves from .54 with the best SFT checkpoint to .597 with supervised GRPO+$R_{\text{think}}$, a +5.7 point improvement. SFT warm-up is critical, as cold-start GRPO drops to a low macro-F1 of .332 on ArMeme. Thinking-length regularization yields the best supervised configuration, confirming that both SFT warm-up and $R_{\text{think}}$ are important for stable GRPO in our setup.

\paragraph{RQ5. Self-supervised RL.}
Self-supervised GRPO improves macro-F1 by +1.5 points on ArMeme, from .597 to .612 (Table~\ref{tab:self_sup_grpo}). Per-class recall improves across all three minority classes, with gains of +1.8 points for propaganda, +3.5 points for not-meme, and +1.8 points for other. On FHM, self-supervised GRPO reduces macro-F1 by +1.0 point (Table~\ref{tab:self_sup_grpo}). This may be due to mismatch in data distribution, as the FHM unlabeled pool is curated from public datasets rather than the FHM distribution. These results suggest that self-supervised GRPO is more effective when the unlabeled and target distributions are aligned and the label space is sufficiently diverse to reduce consensus-based majority-class bias.


\providecommand{\chg}[1]{\textcolor{blue}{#1}}

\section{Discussion}
\label{sec:discussion}
In this section, we reflect on the broader implications of our experimental findings and discuss the main limitations of the current work.


\paragraph{Reward shaping for reasoning stability.}
Our training dynamics analysis (Section~\ref{sec:rl_analysis}, Appendix~\ref{app:training_dynamics}) shows that standard GRPO exhibits a length-shortening bias, particularly on ArMeme. The thresholded lower-bound thinking-length reward $R_{\text{think}}$ mitigates this behavior by penalizing rationales that fall below a minimum reasoning-length threshold, while assigning full reward once the threshold is reached. Thus, $R_{\text{think}}$ encourages sufficient reasoning without incentivizing unnecessary verbosity. This finding suggests that explicit reasoning-length constraints can support more stable RL-based post-training in this setting, as rule-based rewards alone may not be sufficient to maintain rationale quality. More broadly, reward exploitation is a well-documented concern in RL-tuned language models when the reward signal is purely rule-based \citep{singhal2024long, skalse2022defining, firooz2025scaling}. Our results extend this observation to multimodal reasoning tasks and show that even a simple thresholded length penalty can mitigate length-shortening behavior, offering a lightweight alternative to more complex reward-model-based approaches.

\paragraph{Scalability of synthetic supervision.}
Fine-grained labels consistently improve SFT warm-up on both benchmarks, with particularly large gains on ArMeme (Section~\ref{sec:rl_analysis}). Beyond the immediate performance gains, the success of our multi-LLM annotation pipeline (GPT-4.1 + Llama-4-Scout annotators, Gemini-3-Pro as a consolidator) raises a broader question about the scalability of synthetic supervision for content moderation. Our results suggest that multi-source LLM agreement can serve as a useful proxy for human annotation quality in structured labeling tasks, though the extent to which this generalizes to more subjective or culturally nuanced annotation schemes remains an open question. 


\paragraph{Conditions for effective self-supervised RL.}
Self-supervised GRPO succeeds on ArMeme (+1.5\% macro-F1), where the unlabeled data is in-domain and the four-class setting preserves gradient diversity, but degrades on FHM due to distributional mismatch and majority-vote bias in the binary setting. This suggests that consensus-based pseudo-labeling requires \textit{both distribution alignment and sufficient label-space diversity} to provide reliable supervision. This finding has practical implications for deploying RL-based moderation systems. In-domain unlabeled data is abundant on social media platforms, making self-supervised GRPO a viable strategy for continuous model adaptation, provided the label space is sufficiently diverse to avoid consensus-induced label homogenization. However, self-supervised GRPO is sensitive to distribution alignment between unlabeled and target data, and can amplify class biases when distributions diverge, as observed on FHM. These sensitivities suggest that careful data curation and class-aware pseudo-labeling strategies would be needed for robust deployment.

\paragraph{Generative reasoning vs.\ classification.} On ArMeme, our generative reasoning model outperforms the strongest sequence-classification baseline in macro-F1 (0.612 vs.\ 0.594) while also producing explanations. On FHM, our approach also achieves stronger macro-F1 than the sequence-classification baselines. This consistent pattern suggests that generative reasoning models provide more balanced per-class performance than sequence-classification models, while additionally producing interpretable outputs that can support human review. For content moderation in practice, both balanced class performance and actionable explanations are desirable, which favors the generative reasoning approach. 

\paragraph{Limitations.}
We outline several aspects of the present study that point toward natural extensions in future work. Our evaluation focuses on two representative benchmarks, FHM (English) and ArMeme (Arabic), which together span binary and fine-grained labeling across two languages; broader coverage of additional languages, dialects, and culturally specific meme traditions remains an interesting direction. Memes are also a fast-evolving medium, with new templates, visual styles, and cultural references appearing continuously, so periodic re-evaluation on freshly collected data would help track how detection systems generalize over time. We focus on static image-with-text memes; extending the framework to animated GIFs and short-form video memes is left for future work. The thinking-based generation format introduces additional inference-time tokens compared to direct classifiers, which is an inherent trade-off when producing explanations alongside predictions, and deployments with strict latency budgets may benefit from distilled or shorter-CoT variants. Finally, although our self-supervised GRPO reduces reliance on human-labeled data, scaling unlabeled meme collection across many languages still involves non-trivial filtering and curation effort.


\section{Conclusion}
\label{sec_conclusion}

We studied the extent to which GRPO-based training can improve classification performance and reference-based explanation quality in thinking-based MLLMs for meme moderation.
Our RL-based post-training method with a GRPO objective jointly rewards correct classification, structurally valid outputs, and semantically aligned explanations. We additionally constructed extended datasets with distilled CoT rationales and multi-LLM fine-grained propaganda annotations.
Across experiments on both the English Hateful Memes and Arabic ArMeme benchmarks, our approach improves over prior reported results: +2.1\% accuracy on FHM and +7.6\% macro-F1 on ArMeme over prior work. Key findings include: \textit{(i)}~fine-grained supervision and distilled CoT complement RL-based optimization; \textit{(ii)}~multi-LLM annotation pipelines scale fine-grained annotation effectively; \textit{(iii)}~thinking-length regularization mitigates length-shortening bias in rationale generation; and \textit{(iv)}~self-supervised GRPO extracts useful signal from unlabeled in-domain data.
%


\section*{Acknowledgements}
The work of M. B. Kmainasi, A. E. Shahroor and F. Alam was supported by NPRP grant 14C-0916-210015 from the Qatar National Research Fund, part of the Qatar Research Development and Innovation Council (QRDI). The findings reported herein are solely the responsibility of the authors.

\section*{Declaration of generative AI and AI-assisted technologies in the manuscript preparation process}

During the preparation of this work the authors used Claude Opus 4 in order to refine grammar and stylistic phrasing. After using this tool, the authors reviewed and edited the content as needed and take full responsibility for the content of the publication.

\providecommand{\chg}[1]{\textcolor{blue}{#1}}

\appendix

\section{Training Configurations}
\label{app:training_configs}

Tables~\ref{tab:config_unimodal}--\ref{tab:config_multimodal} report the full hyperparameter configurations for all training stages.

\begin{table}[ht!]
\centering
\small
\setlength{\tabcolsep}{4pt}
\caption{Unimodal baseline training configurations.}
\label{tab:config_unimodal}
\begin{tabular}{lcc}
\toprule
\textbf{Parameter} & \textbf{Text} & \textbf{Image} \\
\midrule
Batch Size         & 32  & 512 \\
Learning Rate      & $3{\times}10^{-5}$ & $3{\times}10^{-4}$ \\
Max Seq.\ Length   & 512 & -- \\
Optimizer          & \multicolumn{2}{c}{AdamW} \\
Weight Decay       & 0.01 & 0.01 \\
Epochs             & 10  & 10 \\
Model Selection    & \multicolumn{2}{c}{Validation Acc} \\
Hardware           & \multicolumn{2}{c}{1$\times$ A100} \\
\bottomrule
\end{tabular}
\end{table}

\begin{table}[ht!]
\centering
\small
\setlength{\tabcolsep}{3pt}
\caption{Multimodal training configurations across stages.}
\label{tab:config_multimodal}
\begin{tabular}{lrrr}
\toprule
\textbf{Parameter} & \textbf{SFT} & \textbf{Sup.\ GRPO} & \textbf{SS-GRPO} \\
\midrule
Learning Rate       & $1{\times}10^{-5}$ & $1{\times}10^{-6}$ & $2{\times}10^{-7}$ \\
Batch Size          & 4   & 4   & 4 \\
Grad.\ Accumulation & 1   & 1   & 4 \\
Epochs              & 5   & 5   & 4 \\
Scheduler           & \multicolumn{3}{c}{Cosine (warmup ratio 0.05)} \\
Optimizer           & \multicolumn{3}{c}{AdamW (weight decay 0.1)} \\
Precision           & \multicolumn{3}{c}{bfloat16} \\
Parallelism         & \multicolumn{3}{c}{ZeRO Stage 3} \\
Frozen Modules      & \multicolumn{3}{c}{Vision Encoder} \\
Hardware            & \multicolumn{3}{c}{4$\times$ H200} \\
\midrule
\multicolumn{4}{c}{\textit{GRPO-specific}} \\
\midrule
Candidates ($K$)    & --  & 16  & 16 \\
Temperature         & --  & 1.0 & 1.0 \\
Top-$p$             & --  & 0.85 & 0.85 \\
Top-$k$             & --  & 50  & 50 \\
KL Coefficient      & --  & 0.05 & 0.05 \\
Clip Range ($\epsilon$) & -- & 0.2 & 0.2 \\
Max Length          & --  & 4096 & 4096 \\
Pseudo-labeling     & --  & --  & Majority Vote \\
\bottomrule
\end{tabular}
\end{table}

\section{Qualitative Examples}
\label{app:qualitative}

We present three representative examples (see Figure~\ref{fig:qual_ex1}, \ref{fig:qual_ex2} and \ref{fig:qual_ex3}) illustrating how model predictions improve across training stages. For each example, we show the ground-truth label and explanation alongside the model's prediction before and after a given training stage.

\begin{figure*}[ht!]
    \centering
    \includegraphics[width=\textwidth]{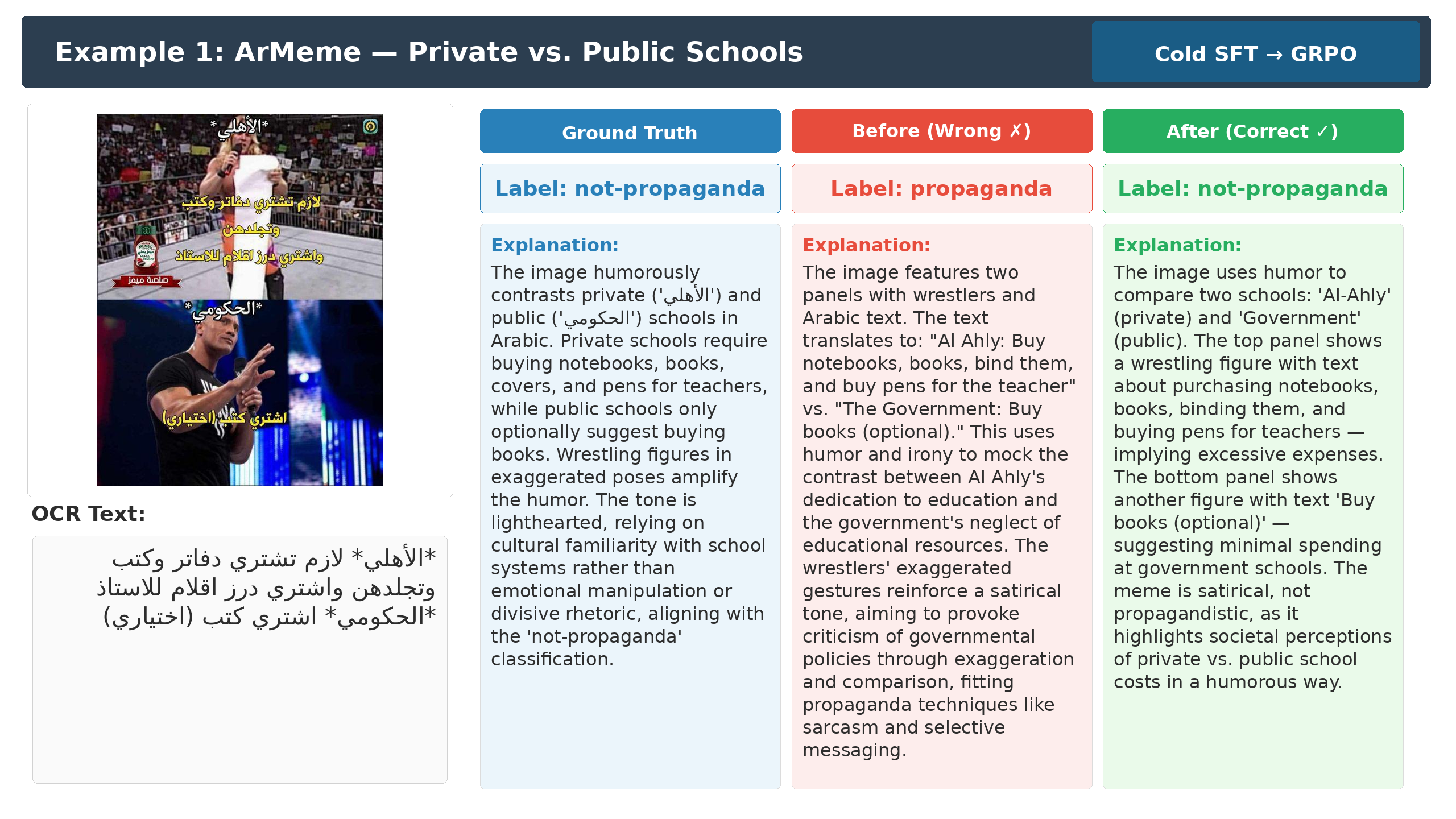}
    \caption{\textbf{Example 1 --- ArMeme: SFT $\to$ GRPO.} The SFT model incorrectly labels a humorous school comparison meme as propaganda, misinterpreting satire as manipulation. After GRPO, the model correctly identifies the lighthearted tone and classifies it as not-propaganda.}
    \label{fig:qual_ex1}
\end{figure*}

\begin{figure*}[ht!]
    \centering
    \includegraphics[width=\textwidth]{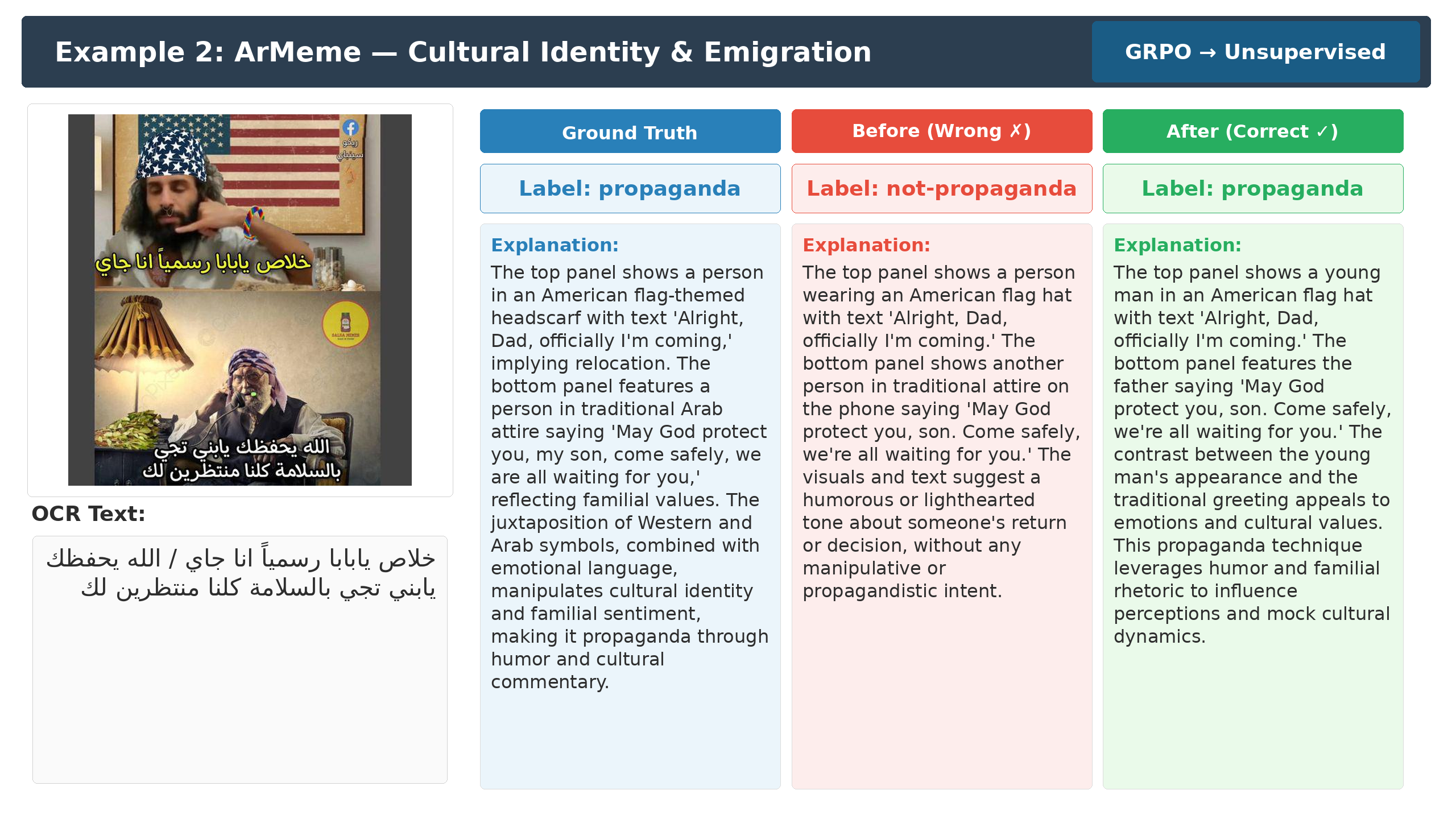}
    \caption{\textbf{Example 2 --- ArMeme: GRPO $\to$ Self-Supervised GRPO.} The supervised model misclassifies a meme that uses cultural identity and emotional language as not-propaganda. After self-supervised GRPO on unlabeled data, the model correctly detects the propaganda technique.}
    \label{fig:qual_ex2}
\end{figure*}

\begin{figure*}[ht!]
    \centering
    \includegraphics[width=\textwidth]{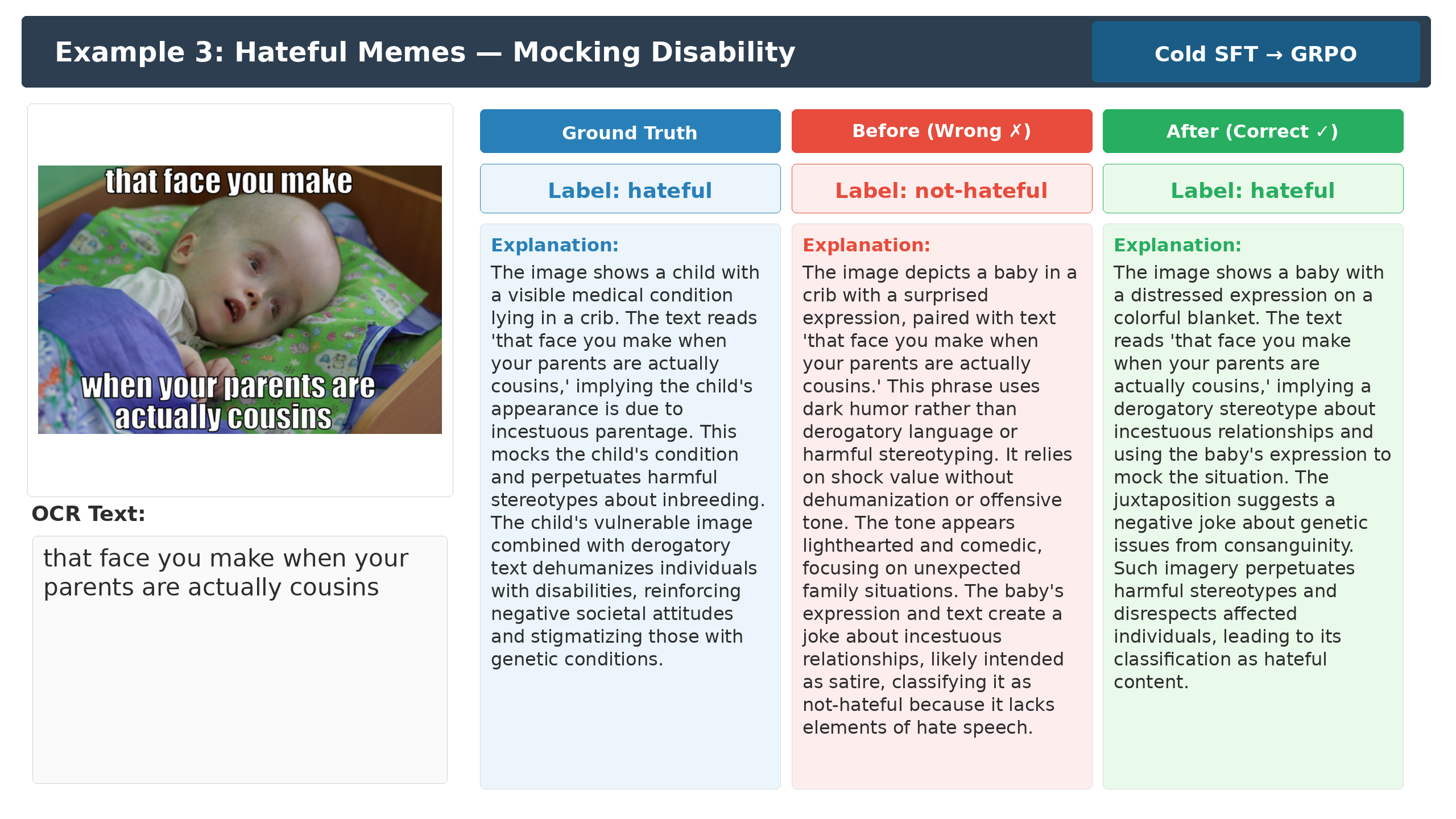}
    \caption{\textbf{Example 3 --- Hateful Memes: SFT $\to$ GRPO.} The SFT model misses implicit hate, interpreting a meme mocking disability through an incest joke as ``dark humor.'' After GRPO, the model's reasoning correctly identifies the dehumanizing stereotype and classifies it as hateful.}
    \label{fig:qual_ex3}
\end{figure*}

\section{GRPO Training Dynamics}
\label{app:training_dynamics}

\begin{figure*}[ht!]
    \centering
    \includegraphics[width=\textwidth]{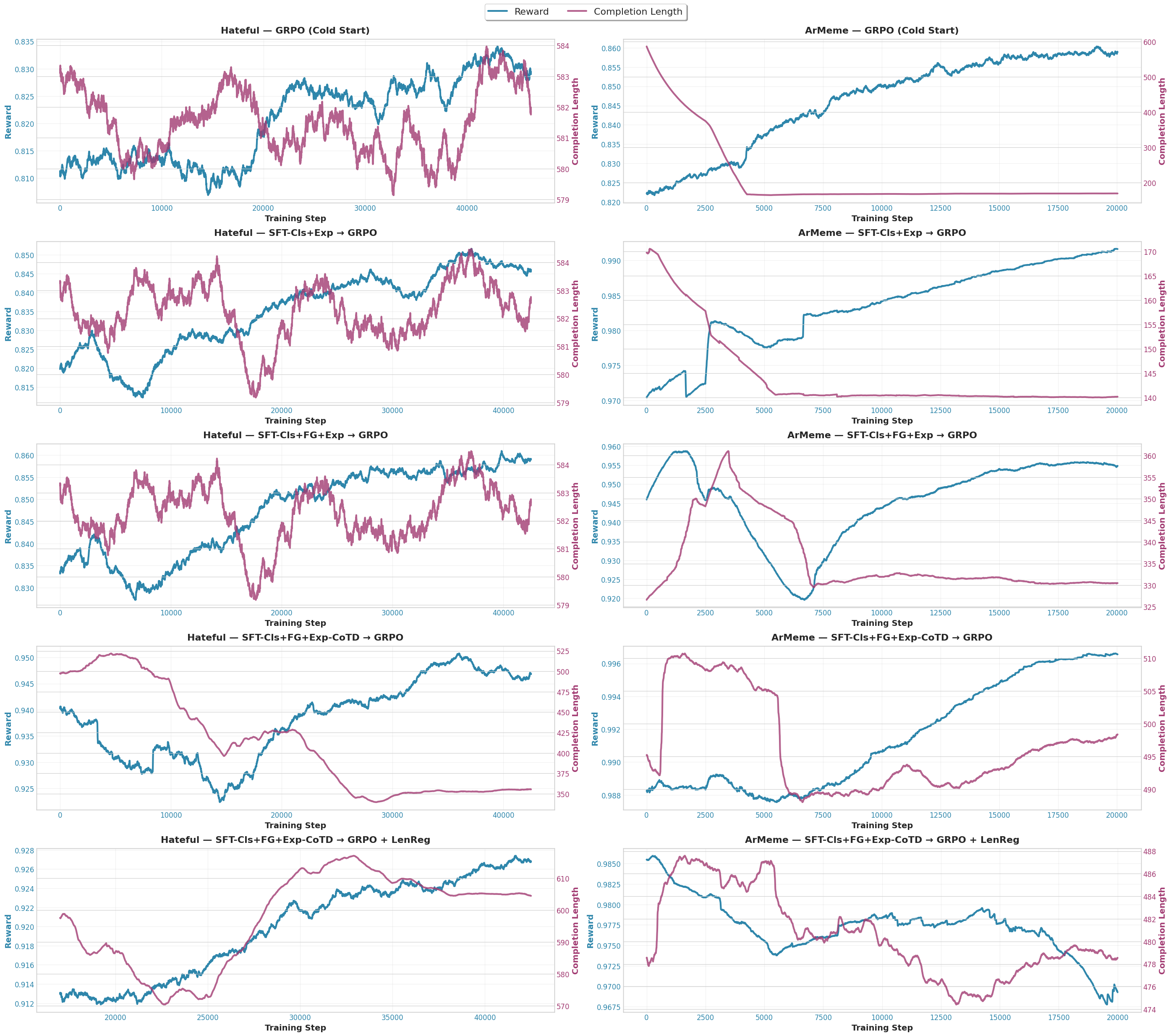}
          \caption{GRPO training dynamics across initialization regimes and datasets, showing mean reward (blue) and completion length (magenta). ArMeme exhibits strong reasoning-length collapse under GRPO, indicating reward hacking through shorter generations. CoT warm-up partially mitigates this effect, while explicit thinking-length regularization ($R_{\text{think}}$) stabilizes reasoning length and prevents collapse. Curves smoothed with a moving average of 1000 steps.}
    \label{fig:reward_dynamics}
\end{figure*}

Figure~\ref{fig:reward_dynamics} compares GRPO reward trajectories and mean completion length across initialization regimes and datasets.
When initialized directly from the pretrained backbone (\emph{cold-start}), GRPO exhibits noisier reward curves with only modest gains and largely stable completion length, indicating that RL alone has difficulty simultaneously learning the required output structure and the task-specific decision boundary. Initializing from SFT (Cls+Exp) yields substantially smoother trajectories and larger gains, suggesting that once the model is aligned to the target schema and supervision distribution, GRPO can focus on improving label correctness and rationale quality. However, on ArMeme, all configurations without thinking-length regularization exhibit a progressive decline in completion length---a symptom of reward hacking through shorter generations. Explicit thinking-length regularization ($R_{\text{think}}$) stabilizes reasoning length and prevents this collapse, as detailed in Section~\ref{sec:rl_analysis}.

\bibliographystyle{elsarticle-harv}

\bibliography{bibliography/custom}

@inproceedings{hasanain-etal-2024-large,
    title = "Large Language Models for Propaganda Span Annotation",
    author = "Hasanain, Maram  and
      Ahmad, Fatema  and
      Alam, Firoj",
    editor = "Al-Onaizan, Yaser  and
      Bansal, Mohit  and
      Chen, Yun-Nung",
    booktitle = "Findings of the Association for Computational Linguistics: EMNLP 2024",
    month = nov,
    year = "2024",
    address = "Miami, Florida, USA",
    publisher = "Association for Computational Linguistics",
    url = "https://aclanthology.org/2024.findings-emnlp.850/",
    doi = "10.18653/v1/2024.findings-emnlp.850",
    pages = "14522--14532"
}

@article{openai2023gpt4,
  title={GPT-4 Technical Report},
  author={{OpenAI}},
  journal={arXiv preprint arXiv:2303.08774},
  year={2023}
}

@article{dubey2024llama,
  title={The {Llama} 3 herd of models},
  author={Dubey, Abhimanyu and Jauhri, Abhinav and Pandey, Abhinav and Kadian, Abhishek and Al-Dahle, Ahmad and Letman, Aiesha and Mathur, Akhil and Schelten, Alan and Yang, Amy and Fan, Angela and others},
  journal={arXiv preprint arXiv:2407.21783},
  year={2024}
}

@article{kamath2025gemma3,
  title         = {Gemma 3 Technical Report},
  author        = {Kamath, Aishwarya and Ferret, Johan and Pathak, Shreya and Vieillard, Nino and Merhej, Ramona and Perrin, Sarah and Matejovicova, Tatiana and Ram{\'e}, Alexandre and Rivi{\`e}re, Morgane and Rouillard, Louis and others},
  journal       = {arXiv preprint arXiv:2503.19786},
  year          = {2025},
  url           = {https://arxiv.org/abs/2503.19786},
}

@article{yang2025qwen3,
  title={Qwen3 technical report},
  author={Yang, An and Li, Anfeng and Yang, Baosong and Zhang, Beichen and Hui, Binyuan and Zheng, Bo and Yu, Bowen and Gao, Chang and Huang, Chengen and Lv, Chenxu and others},
  journal={arXiv preprint arXiv:2505.09388},
  year={2025}
}

@inproceedings{he2016deep,
  title={Deep residual learning for image recognition},
  author={He, Kaiming and Zhang, Xiangyu and Ren, Shaoqing and Sun, Jian},
  booktitle={Proceedings of the IEEE conference on computer vision and pattern recognition},
  pages={770--778},
  year={2016},
  series={CVPR~'16},
  organization={IEEE},
}

@inproceedings{dimitrov-etal-2024-semeval,
  address   = {Mexico City, Mexico},
  author    = {Dimitrov, Dimitar and
               Alam, Firoj and
               Hasanain, Maram and
               Hasnat, Abul and
               Silvestri, Fabrizio and
               Nakov, Preslav and
               Da San Martino, Giovanni},
  editor    = {Ojha, Atul Kr. and
               Do{\u{g}}ru{\"o}z, A. Seza and
               Tayyar Madabushi, Harish and
               Da San Martino, Giovanni and
               Rosenthal, Sara and
               Ros{\'a}, Aiala},
  booktitle = {Proceedings of the 18th International Workshop on Semantic Evaluation},
  series    = {SemEval~'24},
  pages     = {2009--2026},
  publisher = {Association for Computational Linguistics},
  title     = {{SemEval}-2024 Task 4: Multilingual detection of persuasion techniques in memes},
  url       = {https://aclanthology.org/2024.semeval-1.275},
  year      = {2024},
}

@inproceedings{hasanain-etal-2024-araieval,
  address   = {Miami, FL, USA},
  author    = {Hasanain, Maram and
               Hasan, Md. Arid and
               Ahmad, Fatema and
               Suwaileh, Reem and
               Biswas, Md. Rafiul and
               Zaghouani, Wajdi and
               Alam, Firoj},
  editor    = {Habash, Nizar and
               Bouamor, Houda and
               Eskander, Ramy and
               Tomeh, Nadi and
               Abu Farha, Ibrahim and
               Abdelali, Ahmed and
               Touileb, Samia and
               Hamed, Injy and
               Onaizan, Yaser and
               Alhafni, Bashar and
               Antoun, Wissam and
               Khalifa, Salam and
               Haddad, Hatem and
               Zitouni, Imed and
               AlKhamissi, Badr and
               Almatham, Rawan and
               Mrini, Khalil},
  booktitle = {Proceedings of the Second Arabic Natural Language Processing Conference},
  series    = {ArabicNLP~'24},
  pages     = {456--466},
  publisher = {Association for Computational Linguistics},
  title     = {{ArAIEval} shared task: Propagandistic techniques detection in unimodal and multimodal {A}rabic content},
  url       = {https://aclanthology.org/2024.arabicnlp-1.44},
  year      = {2024},
}

@inproceedings{ijcai2022p781,
  address   = {Vienna, Austria},
  author    = {Shivam Sharma and
               Firoj Alam and
               Md. Shad Akhtar and
               Dimitar Dimitrov and
               Giovanni Da San Martino and
               Hamed Firooz and
               Alon Y. Halevy and
               Fabrizio Silvestri and
               Preslav Nakov and
               Tanmoy Chakraborty},
  editor    = {Luc De Raedt},
  booktitle = {Proceedings of the Thirty-First International Joint Conference on Artificial Intelligence},
  series    = {IJCAI~'22},
  pages     = {5597--5606},
  publisher = {International Joint Conferences on Artificial Intelligence Organization},
  title     = {Detecting and understanding harmful memes: {A} survey},
  doi       = {10.24963/IJCAI.2022/781},
  url       = {https://doi.org/10.24963/ijcai.2022/781},
  year      = {2022},
}

@inproceedings{alam-etal-2022-survey,
  address   = {Gyeongju, Republic of Korea},
  author    = {Alam, Firoj and
               Cresci, Stefano and
               Chakraborty, Tanmoy and
               Silvestri, Fabrizio and
               Dimitrov, Dimiter and
               Martino, Giovanni Da San and
               Shaar, Shaden and
               Firooz, Hamed and
               Nakov, Preslav},
  editor    = {Calzolari, Nicoletta and
               Huang, Chu-Ren and
               Kim, Hansaem and
               Pustejovsky, James and
               Wanner, Leo and
               Choi, Key-Sun and
               Ryu, Pum-Mo and
               Chen, Hsin-Hsi and
               Donatelli, Lucia and
               Ji, Heng and
               Kurohashi, Sadao and
               Paggio, Patrizia and
               Xue, Nianwen and
               Kim, Seokhwan and
               Hahm, Younggyun and
               He, Zhong and
               Lee, Tony Kyungil and
               Santus, Enrico and
               Bond, Francis and
               Na, Seung-Hoon},
  booktitle = {Proceedings of the 29th International Conference on Computational Linguistics},
  series    = {COLING~'22},
  pages     = {6625--6643},
  publisher = {International Committee on Computational Linguistics},
  title     = {A survey on multimodal disinformation detection},
  url       = {https://aclanthology.org/2022.coling-1.576},
  year      = {2022},
}

@inproceedings{lu2024mathvista,
  address   = {Vienna, Austria},
  author    = {Pan Lu and
               Hritik Bansal and
               Tony Xia and
               Jiacheng Liu and
               Chunyuan Li and
               Hannaneh Hajishirzi and
               Hao Cheng and
               Kai{-}Wei Chang and
               Michel Galley and
               Jianfeng Gao},
  booktitle = {Proceedings of the International Conference on Learning Representations},
  series    = {ICLR~'24},
  publisher = {OpenReview.net},
  title     = {{MathVista}: Evaluating mathematical reasoning of foundation models in visual contexts},
  url       = {https://openreview.net/forum?id=KUNzEQMWU7},
  year      = {2024},
}

@inproceedings{yue2024mmmu,
  address   = {Seattle, WA, USA},
  author    = {Xiang Yue and
               Yuansheng Ni and
               Tianyu Zheng and
               Kai Zhang and
               Ruoqi Liu and
               Ge Zhang and
               Samuel Stevens and
               Dongfu Jiang and
               Weiming Ren and
               Yuxuan Sun and
               Cong Wei and
               Botao Yu and
               Ruibin Yuan and
               Renliang Sun and
               Ming Yin and
               Boyuan Zheng and
               Zhenzhu Yang and
               Yibo Liu and
               Wenhao Huang and
               Huan Sun and
               Yu Su and
               Wenhu Chen},
  booktitle = {Proceedings of the {IEEE/CVF} Conference on Computer Vision and Pattern Recognition},
  series    = {CVPR~'24},
  doi       = {10.1109/CVPR52733.2024.00913},
  pages     = {9556--9567},
  publisher = {IEEE},
  title     = {{MMMU}: A massive multi-discipline multimodal understanding and reasoning benchmark for expert {AGI}},
  url       = {https://doi.org/10.1109/CVPR52733.2024.00913},
  year      = {2024},
}

@article{openai2023gpt,
  author  = {OpenAI},
  journal = {ArXiv preprint},
  title   = {{GPT-4} technical report},
  url     = {https://arxiv.org/abs/2303.08774},
  volume  = {arXiv:2303.08774},
  year    = {2023},
}

@article{team2025kimi,
  author  = {Team, Kimi and Du, Angang and Yin, Bohong and Xing, Bowei and Qu, Bowen and Wang, Bowen and Chen, Cheng and Zhang, Chenlin and Du, Chenzhuang and Wei, Chu and others},
  journal = {ArXiv preprint},
  title   = {{Kimi-VL} technical report},
  url     = {https://arxiv.org/abs/2504.07491},
  volume  = {arXiv:2504.07491},
  year    = {2025},
}

@article{team2025gemma,
  author  = {Team, Gemma and Kamath, Aishwarya and Ferret, Johan and Pathak, Shreya and Vieillard, Nino and Merhej, Ramona and Perrin, Sarah and Matejovicova, Tatiana and Ram{\'e}, Alexandre and Rivi{\`e}re, Morgane and others},
  journal = {ArXiv preprint},
  title   = {{Gemma}~3 technical report},
  url     = {https://arxiv.org/abs/2503.19786},
  volume  = {arXiv:2503.19786},
  year    = {2025},
}

@article{comanici2025gemini,
  author  = {Comanici, Gheorghe and Bieber, Eric and Schaekermann, Mike and Pasupat, Ice and Sachdeva, Noveen and Dhillon, Inderjit and Blistein, Marcel and Ram, Ori and Zhang, Dan and Rosen, Evan and others},
  journal = {ArXiv preprint},
  title   = {{Gemini}~2.5: Pushing the frontier with advanced reasoning, multimodality, long context, and next generation agentic capabilities},
  url     = {https://arxiv.org/abs/2507.06261},
  volume  = {arXiv:2507.06261},
  year    = {2025},
}

@article{latif2025openaiO1,
  author  = {Latif, Ehsan and Zhou, Yifan and Guo, Shuchen and Gao, Yizhu and Shi, Lehong and Nyaaba, Matthew and Bewerdorff, Arne and Yang, Xiantong and Zhai, Xiaoming and others},
  doi     = {10.1038/s41598-025-33629-9},
  journal = {Scientific Reports},
  title   = {Comparative evaluation of {OpenAI} {o1} and human performance in higher order cognition},
  url     = {https://doi.org/10.1038/s41598-025-33629-9},
  year    = {2025},
}

@article{mei2025expo,
  author  = {Mei, Jingbiao and Sun, Mingsheng and Chen, Jinghong and Qin, Pengda and Li, Yuhong and Chen, Da and Byrne, Bill},
  journal = {ArXiv preprint},
  title   = {{ExPO-HM}: Learning to explain-then-detect for hateful meme detection},
  url     = {https://arxiv.org/abs/2510.08630},
  volume  = {arXiv:2510.08630},
  year    = {2025},
}

@article{bai2025qwen3vltechnicalreport,
  author  = {Bai, Shuai and Cai, Yuxuan and Chen, Ruizhe and Chen, Keqin and Chen, Xionghui and Cheng, Zesen and Deng, Lianghao and Ding, Wei and Gao, Chang and Ge, Chunjiang and Ge, Wenbin and Guo, Zhifang and Huang, Qidong and Huang, Jie and Huang, Fei and Hui, Binyuan and Jiang, Shutong and Li, Zhaohai and Li, Mingsheng and Li, Mei and Li, Kaixin and Lin, Zicheng and Lin, Junyang and Liu, Xuejing and Liu, Jiawei and Liu, Chenglong and Liu, Yang and Liu, Dayiheng and Liu, Shixuan and Lu, Dunjie and Luo, Ruilin and Lv, Chenxu and Men, Rui and Meng, Lingchen and Ren, Xuancheng and Ren, Xingzhang and Song, Sibo and Sun, Yuchong and Tang, Jun and Tu, Jianhong and Wan, Jianqiang and Wang, Peng and Wang, Pengfei and Wang, Qiuyue and Wang, Yuxuan and Xie, Tianbao and Xu, Yiheng and Xu, Haiyang and Xu, Jin and Yang, Zhibo and Yang, Mingkun and Yang, Jianxin and Yang, An and Yu, Bowen and Zhang, Fei and Zhang, Hang and Zhang, Xi and Zheng, Bo and Zhong, Humen and Zhou, Jingren and Zhou, Fan and Zhou, Jing and Zhu, Yuanzhi and Zhu, Ke},
  journal = {ArXiv preprint},
  title   = {{Qwen3-VL} technical report},
  url     = {https://arxiv.org/abs/2511.21631},
  volume  = {arXiv:2511.21631},
  year    = {2025},
}

@article{li2026qwen3,
  author  = {Li, Mingxin and Zhang, Yanzhao and Long, Dingkun and Chen, Keqin and Song, Sibo and Bai, Shuai and Yang, Zhibo and Xie, Pengjun and Yang, An and Liu, Dayiheng and others},
  journal = {ArXiv preprint},
  title   = {{Qwen3-VL}-Embedding and {Qwen3-VL}-Reranker: A unified framework for state-of-the-art multimodal retrieval and ranking},
  url     = {https://arxiv.org/abs/2601.04720},
  volume  = {arXiv:2601.04720},
  year    = {2026},
}

@article{pandiani2025toxic,
  author    = {Pandiani, Delfina S Martinez and Sang, Erik Tjong Kim and Ceolin, Davide},
  doi       = {10.1016/j.osnem.2025.100317},
  journal   = {Online Social Networks and Media},
  pages     = {100317},
  publisher = {Elsevier},
  title     = {`{Toxic}' memes: A survey of computational perspectives on the detection and explanation of meme toxicities},
  url       = {https://doi.org/10.1016/j.osnem.2025.100317},
  volume    = {47},
  year      = {2025},
}

@article{alafnan2025role,
  author    = {AlAfnan, Mohammad Awad},
  doi       = {10.11114/smc.v13i2.7482},
  journal   = {Studies in Media and Communication},
  number    = {2},
  pages     = {1--10},
  publisher = {Redfame Publishing},
  title     = {The role of memes in shaping political discourse on social media},
  url       = {https://doi.org/10.11114/smc.v13i2.7482},
  volume    = {13},
  year      = {2025},
}

@article{schmid2025humorous,
  author    = {Schmid, Ursula Kristin},
  doi       = {10.1177/14614448231198169},
  journal   = {New Media \& Society},
  number    = {3},
  pages     = {1588--1606},
  publisher = {Sage Publications},
  title     = {Humorous hate speech on social media: A mixed-methods investigation of users' perceptions and processing of hateful memes},
  url       = {https://doi.org/10.1177/14614448231198169},
  volume    = {27},
  year      = {2025},
}

@article{mihuailescu2024never,
  author    = {Mih{\u{a}}ilescu, Mihaela-Georgiana},
  doi       = {10.1177/20563051241296256},
  journal   = {Social Media {+} Society},
  number    = {4},
  pages     = {20563051241296256},
  publisher = {Sage Publications},
  title     = {Never mess with the ``memers'': How meme creators are redefining contemporary politics},
  url       = {https://doi.org/10.1177/20563051241296256},
  volume    = {10},
  year      = {2024},
}

@inproceedings{amalia2018meme,
  address   = {Palembang, Indonesia},
  author    = {Amalia, Amalia and Sharif, Amer and Haisar, Fikri and Gunawan, Dani and Nasution, Benny B},
  booktitle = {Proceedings of the Third International Conference on Informatics and Computing},
  series    = {ICIC~'18},
  doi       = {10.1109/IAC.2018.8780410},
  pages     = {1--5},
  publisher = {IEEE},
  title     = {Meme opinion categorization by using optical character recognition ({OCR}) and {Na{\"i}ve} {Bayes} algorithm},
  url       = {https://doi.org/10.1109/IAC.2018.8780410},
  year      = {2018},
}

@inproceedings{boinepelli2020sis,
  address   = {Barcelona (online)},
  author    = {Boinepelli, Sravani and
               Shrivastava, Manish and
               Varma, Vasudeva},
  editor    = {Herbelot, Aurelie and
               Zhu, Xiaodan and
               Palmer, Alexis and
               Schneider, Nathan and
               May, Jonathan and
               Shutova, Ekaterina},
  booktitle = {Proceedings of the Fourteenth Workshop on Semantic Evaluation},
  series    = {SemEval~'20},
  doi       = {10.18653/v1/2020.semeval-1.157},
  pages     = {1190--1194},
  publisher = {International Committee for Computational Linguistics},
  title     = {{SIS}@{IIITH} at {SemEval}-2020 Task 8: An overview of simple text classification methods for meme analysis},
  url       = {https://aclanthology.org/2020.semeval-1.157},
  year      = {2020},
}

@inproceedings{shrestha2020nlp_uiowa,
  address   = {Barcelona (online)},
  author    = {Shrestha, Ingroj and
               Rusert, Jonathan},
  editor    = {Herbelot, Aurelie and
               Zhu, Xiaodan and
               Palmer, Alexis and
               Schneider, Nathan and
               May, Jonathan and
               Shutova, Ekaterina},
  booktitle = {Proceedings of the Fourteenth Workshop on Semantic Evaluation},
  series    = {SemEval~'20},
  doi       = {10.18653/v1/2020.semeval-1.113},
  pages     = {891--900},
  publisher = {International Committee for Computational Linguistics},
  title     = {{NLP\_UIOWA} at {SemEval}-2020 Task 8: You're not the only one cursed with knowledge --- multi branch model memotion analysis},
  url       = {https://aclanthology.org/2020.semeval-1.113},
  year      = {2020},
}

@inproceedings{kiela2020hateful,
  address   = {Virtual},
  author    = {Douwe Kiela and
               Hamed Firooz and
               Aravind Mohan and
               Vedanuj Goswami and
               Amanpreet Singh and
               Pratik Ringshia and
               Davide Testuggine},
  editor    = {Hugo Larochelle and
               Marc'Aurelio Ranzato and
               Raia Hadsell and
               Maria{-}Florina Balcan and
               Hsuan{-}Tien Lin},
  booktitle = {Advances in Neural Information Processing Systems},
  series    = {NeurIPS~'20},
  title     = {The hateful memes challenge: Detecting hate speech in multimodal memes},
  url       = {https://proceedings.neurips.cc/paper/2020/hash/1b84c4cee2b8b3d823b30e2d604b1878-Abstract.html},
  year      = {2020},
}

@inproceedings{lu2025having,
  address   = {Padua, Italy},
  author    = {Lu, Junyu and Xu, Bo and Zhang, Xiaokun and Zhu, Haohao and Wang, Kaichun and Yang, Liang and Lin, Hongfei},
  booktitle = {Proceedings of the 48th International {ACM} {SIGIR} Conference on Research and Development in Information Retrieval},
  series    = {SIGIR~'25},
  doi       = {10.1145/3726302.3730014},
  pages     = {559--569},
  publisher = {ACM},
  title     = {Is having rationales enough? {R}ethinking knowledge enhancement for multimodal hateful meme detection},
  url       = {https://doi.org/10.1145/3726302.3730014},
  year      = {2025},
}

@inproceedings{cao2024modularized,
  address   = {Singapore},
  author    = {Rui Cao and
               Roy Ka{-}Wei Lee and
               Jing Jiang},
  editor    = {Tat{-}Seng Chua and
               Chong{-}Wah Ngo and
               Ravi Kumar and
               Hady W. Lauw and
               Roy Ka{-}Wei Lee},
  booktitle = {Proceedings of the {ACM} Web Conference 2024},
  series    = {WWW~'24},
  doi       = {10.1145/3589334.3648145},
  pages     = {4575--4584},
  publisher = {ACM},
  title     = {Modularized networks for few-shot hateful meme detection},
  url       = {https://doi.org/10.1145/3589334.3648145},
  year      = {2024},
}

@inproceedings{heebridging,
  address   = {Miami, FL, USA},
  author    = {Hee, Ming Shan and Kumaresan, Aditi and Lee, Roy Ka-Wei},
  booktitle = {Proceedings of the 2024 Conference on Empirical Methods in Natural Language Processing},
  series    = {EMNLP~'24},
  doi       = {10.18653/v1/2024.emnlp-main.445},
  publisher = {Association for Computational Linguistics},
  title     = {Bridging modalities: Enhancing cross-modality hate speech detection with few-shot in-context learning},
  url       = {https://doi.org/10.18653/v1/2024.emnlp-main.445},
  year      = {2024},
}

@article{liu2025mind,
  author  = {Liu, Ziyan and Fan, Chunxiao and Lou, Haoran and Wu, Yuexin and Deng, Kaiwei},
  journal = {ArXiv preprint},
  title   = {{MIND}: A multi-agent framework for zero-shot harmful meme detection},
  url     = {https://arxiv.org/abs/2507.06908},
  volume  = {arXiv:2507.06908},
  year    = {2025},
}

@inproceedings{kmainasi2025memeintel,
  address   = {Suzhou, China},
  author    = {Kmainasi, Mohamed Bayan and
               Hasnat, Abul and
               Hasan, Md Arid and
               Shahroor, Ali Ezzat and
               Alam, Firoj},
  editor    = {Christodoulopoulos, Christos and
               Chakraborty, Tanmoy and
               Rose, Carolyn and
               Peng, Violet},
  booktitle = {Proceedings of the 2025 Conference on Empirical Methods in Natural Language Processing},
  series    = {EMNLP~'25},
  doi       = {10.18653/v1/2025.emnlp-main.1539},
  isbn      = {979-8-89176-332-6},
  pages     = {30251--30267},
  publisher = {Association for Computational Linguistics},
  title     = {{MemeIntel}: Explainable detection of propagandistic and hateful memes},
  url       = {https://aclanthology.org/2025.emnlp-main.1539/},
  year      = {2025},
}

@inproceedings{wu2024visionllm,
  address   = {Vancouver, BC, Canada},
  author    = {Jiannan Wu and
               Muyan Zhong and
               Sen Xing and
               Zeqiang Lai and
               Zhaoyang Liu and
               Zhe Chen and
               Wenhai Wang and
               Xizhou Zhu and
               Lewei Lu and
               Tong Lu and
               Ping Luo and
               Yu Qiao and
               Jifeng Dai},
  editor    = {Amir Globersons and
               Lester Mackey and
               Danielle Belgrave and
               Angela Fan and
               Ulrich Paquet and
               Jakub M. Tomczak and
               Cheng Zhang},
  booktitle = {Advances in Neural Information Processing Systems},
  series    = {NeurIPS~'24},
  title     = {{VisionLLM}~v2: An end-to-end generalist multimodal large language model for hundreds of vision-language tasks},
  url       = {http://papers.nips.cc/paper_files/paper/2024/hash/81a60d18e010b27b36cd465c6604b915-Abstract-Conference.html},
  year      = {2024},
}

@inproceedings{sun2025multi,
  address   = {Tucson, AZ, USA},
  author    = {Sun, Li and Ahuja, Chaitanya and Chen, Peng and D'Zmura, Matt and Batmanghelich, Kayhan and Bontrager, Philip},
  booktitle = {Proceedings of the {IEEE/CVF} Winter Conference on Applications of Computer Vision},
  series    = {WACV~'25},
  pages     = {8617--8626},
  publisher = {IEEE},
  title     = {Multi-modal large language models are effective vision learners},
  url       = {https://openaccess.thecvf.com/content/WACV2025/html/Sun_Multi-Modal_Large_Language_Models_are_Effective_Vision_Learners_WACV_2025_paper.html},
  year      = {2025},
}

@article{wei2025sftsecondrlupt,
  author  = {Lai Wei and Yuting Li and Chen Wang and Yue Wang and Linghe Kong and Weiran Huang and Lichao Sun},
  journal = {ArXiv preprint},
  title   = {First {SFT}, second {RL}, third {UPT}: Continual improving multi-modal {LLM} reasoning via unsupervised post-training},
  url     = {https://arxiv.org/abs/2505.22453},
  volume  = {arXiv:2505.22453},
  year    = {2025},
}

@inproceedings{yu2024rlhf,
  address   = {Seattle, WA, USA},
  author    = {Tianyu Yu and
               Yuan Yao and
               Haoye Zhang and
               Taiwen He and
               Yifeng Han and
               Ganqu Cui and
               Jinyi Hu and
               Zhiyuan Liu and
               Hai{-}Tao Zheng and
               Maosong Sun},
  booktitle = {Proceedings of the {IEEE/CVF} Conference on Computer Vision and Pattern Recognition},
  series    = {CVPR~'24},
  doi       = {10.1109/CVPR52733.2024.01310},
  pages     = {13807--13816},
  publisher = {IEEE},
  title     = {{RLHF-V}: Towards trustworthy {MLLMs} via behavior alignment from fine-grained correctional human feedback},
  url       = {https://doi.org/10.1109/CVPR52733.2024.01310},
  year      = {2024},
}

@article{deng2025openvlthinker,
  author  = {Deng, Yihe and Bansal, Hritik and Yin, Fan and Peng, Nanyun and Wang, Wei and Chang, Kai-Wei},
  journal = {ArXiv preprint},
  title   = {{OpenVLThinker}: An early exploration to complex vision-language reasoning via iterative self-improvement},
  url     = {https://arxiv.org/abs/2503.17352},
  volume  = {arXiv:2503.17352},
  year    = {2025},
}

@article{kaelbling1996reinforcementlearningsurvey,
  author  = {L. P. Kaelbling and M. L. Littman and A. W. Moore},
  journal = {Journal of Artificial Intelligence Research},
  pages   = {237--285},
  title   = {Reinforcement learning: A survey},
  volume  = {4},
  year    = {1996},
}

@article{wu2025sailing,
  author  = {Wu, Xiaobao},
  journal = {ArXiv preprint},
  title   = {Sailing by the stars: A survey on reward models and learning strategies for learning from rewards},
  url     = {https://arxiv.org/abs/2505.02686},
  volume  = {arXiv:2505.02686},
  year    = {2025},
}

@inproceedings{ouyang2022training,
  address   = {New Orleans, LA, USA},
  author    = {Long Ouyang and
               Jeffrey Wu and
               Xu Jiang and
               Diogo Almeida and
               Carroll L. Wainwright and
               Pamela Mishkin and
               Chong Zhang and
               Sandhini Agarwal and
               Katarina Slama and
               Alex Ray and
               John Schulman and
               Jacob Hilton and
               Fraser Kelton and
               Luke Miller and
               Maddie Simens and
               Amanda Askell and
               Peter Welinder and
               Paul F. Christiano and
               Jan Leike and
               Ryan Lowe},
  booktitle = {Advances in Neural Information Processing Systems},
  series    = {NeurIPS~'22},
  title     = {Training language models to follow instructions with human feedback},
  url       = {http://papers.nips.cc/paper_files/paper/2022/hash/b1efde53be364a73914f58805a001731-Abstract-Conference.html},
  year      = {2022},
}

@article{schulman2017proximal,
  author  = {Schulman, John and Wolski, Filip and Dhariwal, Prafulla and Radford, Alec and Klimov, Oleg},
  journal = {ArXiv preprint},
  title   = {Proximal policy optimization algorithms},
  url     = {https://arxiv.org/abs/1707.06347},
  volume  = {arXiv:1707.06347},
  year    = {2017},
}

@inproceedings{rafailov2023direct,
  address   = {New Orleans, LA, USA},
  author    = {Rafael Rafailov and
               Archit Sharma and
               Eric Mitchell and
               Christopher D. Manning and
               Stefano Ermon and
               Chelsea Finn},
  booktitle = {Advances in Neural Information Processing Systems},
  series    = {NeurIPS~'23},
  title     = {Direct preference optimization: Your language model is secretly a reward model},
  url       = {http://papers.nips.cc/paper_files/paper/2023/hash/a85b405ed65c6477a4fe8302b5e06ce7-Abstract-Conference.html},
  year      = {2023},
}

@article{deepseekai2025deepseekr1incentivizingreasoningcapability,
  author  = {DeepSeek-{AI} and Daya Guo and Dejian Yang and Haowei Zhang and Junxiao Song and Ruoyu Zhang and Runxin Xu and Qihao Zhu and Shirong Ma and Peiyi Wang and Xiao Bi and Xiaokang Zhang and Xingkai Yu and Yu Wu and Z. F. Wu and Zhibin Gou and Zhihong Shao and Zhuoshu Li and Ziyi Gao and Aixin Liu and Bing Xue and Bingxuan Wang and Bochao Wu and Bei Feng and Chengda Lu and Chenggang Zhao and Chengqi Deng and Chenyu Zhang and Chong Ruan and Damai Dai and Deli Chen and Dongjie Ji and Erhang Li and Fangyun Lin and Fucong Dai and Fuli Luo and Guangbo Hao and Guanting Chen and Guowei Li and H. Zhang and Han Bao and Hanwei Xu and Haocheng Wang and Honghui Ding and Huajian Xin and Huazuo Gao and Hui Qu and Hui Li and Jianzhong Guo and Jiashi Li and Jiawei Wang and Jingchang Chen and Jingyang Yuan and Junjie Qiu and Junlong Li and J. L. Cai and Jiaqi Ni and Jian Liang and Jin Chen and Kai Dong and Kai Hu and Kaige Gao and Kang Guan and Kexin Huang and Kuai Yu and Lean Wang and Lecong Zhang and Liang Zhao and Litong Wang and Liyue Zhang and Lei Xu and Leyi Xia and Mingchuan Zhang and Minghua Zhang and Minghui Tang and Meng Li and Miaojun Wang and Mingming Li and Ning Tian and Panpan Huang and Peng Zhang and Qiancheng Wang and Qinyu Chen and Qiushi Du and Ruiqi Ge and Ruisong Zhang and Ruizhe Pan and Runji Wang and R. J. Chen and R. L. Jin and Ruyi Chen and Shanghao Lu and Shangyan Zhou and Shanhuang Chen and Shengfeng Ye and Shiyu Wang and Shuiping Yu and Shunfeng Zhou and Shuting Pan and S. S. Li and Shuang Zhou and Shaoqing Wu and Tao Yun and Tian Pei and Tianyu Sun and T. Wang and Wangding Zeng and Wanjia Zhao and Wen Liu and Wenfeng Liang and Wenjun Gao and Wenqin Yu and Wentao Zhang and W. L. Xiao and Wei An and Xiaodong Liu and Xiaohan Wang and Xiaokang Chen and Xiaotao Nie and Xin Cheng and Xin Liu and Xin Xie and Xingchao Liu and Xinyu Yang and Xinyuan Li and Xuecheng Su and Xuheng Lin and X. Q. Li and Xiangyue Jin and Xiaojin Shen and Xiaosha Chen and Xiaowen Sun and Xiaoxiang Wang and Xinnan Song and Xinyi Zhou and Xianzu Wang and Xinxia Shan and Y. K. Li and Y. Q. Wang and Y. X. Wei and Yang Zhang and Yanhong Xu and Yao Li and Yao Zhao and Yaofeng Sun and Yaohui Wang and Yi Yu and Yichao Zhang and Yifan Shi and Yiliang Xiong and Ying He and Yishi Piao and Yisong Wang and Yixuan Tan and Yiyang Ma and Yiyuan Liu and Yongqiang Guo and Yuan Ou and Yuduan Wang and Yue Gong and Yuheng Zou and Yujia He and Yunfan Xiong and Yuxiang Luo and Yuxiang You and Yuxuan Liu and Yuyang Zhou and Y. X. Zhu and Yanping Huang and Yaohui Li and Yi Zheng and Yuchen Zhu and Yunxian Ma and Ying Tang and Yukun Zha and Yuting Yan and Z. Z. Ren and Zehui Ren and Zhangli Sha and Zhe Fu and Zhean Xu and Zhenda Xie and Zhengyan Zhang and Zhewen Hao and Zhicheng Ma and Zhigang Yan and Zhiyu Wu and Zihui Gu and Zijia Zhu and Zijun Liu and Zilin Li and Ziwei Xie and Ziyang Song and Zizheng Pan and Zhen Huang and Zhipeng Xu and Zhongyu Zhang and Zhen Zhang},
  journal = {ArXiv preprint},
  title   = {{DeepSeek-R1}: Incentivizing reasoning capability in {LLMs} via reinforcement learning},
  url     = {https://arxiv.org/abs/2501.12948},
  volume  = {arXiv:2501.12948},
  year    = {2025},
}

@article{shao2024deepseekmathpushinglimitsmathematical,
  author  = {Zhihong Shao and Peiyi Wang and Qihao Zhu and Runxin Xu and Junxiao Song and Xiao Bi and Haowei Zhang and Mingchuan Zhang and Y. K. Li and Y. Wu and Daya Guo},
  journal = {ArXiv preprint},
  title   = {{DeepSeekMath}: Pushing the limits of mathematical reasoning in open language models},
  url     = {https://arxiv.org/abs/2402.03300},
  volume  = {arXiv:2402.03300},
  year    = {2024},
}

@article{tan2025gtpo,
  author  = {Tan, Hongze and Pan, Jianfei},
  journal = {ArXiv preprint},
  title   = {{GTPO} and {GRPO-S}: Token- and sequence-level reward shaping with policy entropy},
  url     = {https://arxiv.org/abs/2508.04349},
  volume  = {arXiv:2508.04349},
  year    = {2025},
}

@inproceedings{ranaldi2025multilingual,
  address   = {Albuquerque, NM, USA},
  author    = {Ranaldi, Leonardo and Pucci, Giulia},
  booktitle = {Proceedings of the 2025 Conference of the Nations of the Americas Chapter of the Association for Computational Linguistics: Human Language Technologies (Volume 1: Long Papers)},
  series    = {NAACL~'25},
  doi       = {10.18653/v1/2025.naacl-long.577},
  pages     = {11566--11582},
  publisher = {Association for Computational Linguistics},
  title     = {Multilingual reasoning via self-training},
  url       = {https://doi.org/10.18653/v1/2025.naacl-long.577},
  year      = {2025},
}

@inproceedings{chen2025predicate,
  address   = {Suzhou, China},
  author    = {Chen, Jiajun and Tam, Yik-Cheung},
  booktitle = {Proceedings of the 2025 Conference on Empirical Methods in Natural Language Processing},
  series    = {EMNLP~'25},
  doi       = {10.18653/v1/2025.emnlp-main.462},
  pages     = {9097--9110},
  publisher = {Association for Computational Linguistics},
  title     = {Predicate-guided generation for mathematical reasoning},
  url       = {https://doi.org/10.18653/v1/2025.emnlp-main.462},
  year      = {2025},
}

@inproceedings{pramanick-etal-2021-momenta-multimodal,
  address   = {Punta Cana, Dominican Republic},
  author    = {Pramanick, Shraman and
               Sharma, Shivam and
               Dimitrov, Dimitar and
               Akhtar, Md. Shad and
               Nakov, Preslav and
               Chakraborty, Tanmoy},
  editor    = {Moens, Marie-Francine and
               Huang, Xuanjing and
               Specia, Lucia and
               Yih, Scott Wen-tau},
  booktitle = {Findings of the Association for Computational Linguistics: {EMNLP} 2021},
  series    = {Findings~'21},
  doi       = {10.18653/v1/2021.findings-emnlp.379},
  pages     = {4439--4455},
  publisher = {Association for Computational Linguistics},
  title     = {{MOMENTA}: A multimodal framework for detecting harmful memes and their targets},
  url       = {https://aclanthology.org/2021.findings-emnlp.379},
  year      = {2021},
}

@inproceedings{cao2023pro,
  address   = {Ottawa, ON, Canada},
  author    = {Cao, Rui and Hee, Ming Shan and Kuek, Adriel and Chong, Wen-Haw and Lee, Roy Ka-Wei and Jiang, Jing},
  booktitle = {Proceedings of the 31st {ACM} International Conference on Multimedia},
  series    = {MM~'23},
  doi       = {10.1145/3581783.3612498},
  pages     = {5244--5252},
  publisher = {ACM},
  title     = {{Pro-Cap}: Leveraging a frozen vision-language model for hateful meme detection},
  url       = {https://doi.org/10.1145/3581783.3612498},
  year      = {2023},
}

@inproceedings{mei-etal-2024-improving,
  address   = {Bangkok, Thailand},
  author    = {Mei, Jingbiao and
               Chen, Jinghong and
               Lin, Weizhe and
               Byrne, Bill and
               Tomalin, Marcus},
  editor    = {Ku, Lun-Wei and
               Martins, Andre and
               Srikumar, Vivek},
  booktitle = {Proceedings of the 62nd Annual Meeting of the Association for Computational Linguistics (Volume 1: Long Papers)},
  series    = {ACL~'24},
  doi       = {10.18653/v1/2024.acl-long.291},
  pages     = {5333--5347},
  publisher = {Association for Computational Linguistics},
  title     = {Improving hateful meme detection through retrieval-guided contrastive learning},
  url       = {https://aclanthology.org/2024.acl-long.291/},
  year      = {2024},
}

@inproceedings{fersini-etal-2022-semeval,
  address   = {Seattle, WA, USA},
  author    = {Fersini, Elisabetta and
               Gasparini, Francesca and
               Rizzi, Giulia and
               Saibene, Aurora and
               Chulvi, Berta and
               Rosso, Paolo and
               Lees, Alyssa and
               Sorensen, Jeffrey},
  editor    = {Emerson, Guy and
               Schluter, Natalie and
               Stanovsky, Gabriel and
               Kumar, Ritesh and
               Palmer, Alexis and
               Schneider, Nathan and
               Singh, Siddharth and
               Ratan, Shyam},
  booktitle = {Proceedings of the 16th International Workshop on Semantic Evaluation},
  series    = {SemEval~'22},
  doi       = {10.18653/v1/2022.semeval-1.74},
  pages     = {533--549},
  publisher = {Association for Computational Linguistics},
  title     = {{SemEval}-2022 Task 5: Multimedia automatic misogyny identification},
  url       = {https://aclanthology.org/2022.semeval-1.74},
  year      = {2022},
}

@inproceedings{alam-etal-2024-armeme,
  address   = {Miami, FL, USA},
  author    = {Alam, Firoj and
               Hasnat, Abul and
               Ahmad, Fatema and
               Hasan, Md. Arid and
               Hasanain, Maram},
  editor    = {Al-Onaizan, Yaser and
               Bansal, Mohit and
               Chen, Yun-Nung},
  booktitle = {Proceedings of the 2024 Conference on Empirical Methods in Natural Language Processing},
  series    = {EMNLP~'24},
  doi       = {10.18653/v1/2024.emnlp-main.1173},
  pages     = {21071--21090},
  publisher = {Association for Computational Linguistics},
  title     = {{ArMeme}: Propagandistic content in {A}rabic memes},
  url       = {https://aclanthology.org/2024.emnlp-main.1173/},
  year      = {2024},
}

@inproceedings{mathias-etal-2021-findings,
  address   = {Online},
  author    = {Mathias, Lambert and
               Nie, Shaoliang and
               Mostafazadeh Davani, Aida and
               Kiela, Douwe and
               Prabhakaran, Vinodkumar and
               Vidgen, Bertie and
               Waseem, Zeerak},
  editor    = {Mostafazadeh Davani, Aida and
               Kiela, Douwe and
               Lambert, Mathias and
               Vidgen, Bertie and
               Prabhakaran, Vinodkumar and
               Waseem, Zeerak},
  booktitle = {Proceedings of the 5th Workshop on Online Abuse and Harms ({WOAH} 2021)},
  series    = {WOAH~'21},
  doi       = {10.18653/v1/2021.woah-1.21},
  pages     = {201--206},
  publisher = {Association for Computational Linguistics},
  title     = {Findings of the {WOAH} 5 shared task on fine grained hateful memes detection},
  url       = {https://aclanthology.org/2021.woah-1.21},
  year      = {2021},
}

@article{grasso2024kermit,
  author    = {Grasso, Biagio and La Gatta, Valerio and Moscato, Vincenzo and Sperl{\`\i}, Giancarlo},
  doi       = {10.1016/j.inffus.2024.102269},
  journal   = {Information Fusion},
  pages     = {102269},
  publisher = {Elsevier},
  title     = {{KERMIT}: Knowledge-empowered model in harmful meme detection},
  url       = {https://doi.org/10.1016/j.inffus.2024.102269},
  volume    = {106},
  year      = {2024},
}

@inproceedings{lin2024towards,
  address   = {Singapore},
  author    = {Hongzhan Lin and
               Ziyang Luo and
               Wei Gao and
               Jing Ma and
               Bo Wang and
               Ruichao Yang},
  editor    = {Tat{-}Seng Chua and
               Chong{-}Wah Ngo and
               Ravi Kumar and
               Hady W. Lauw and
               Roy Ka{-}Wei Lee},
  booktitle = {Proceedings of the {ACM} Web Conference 2024},
  series    = {WWW~'24},
  doi       = {10.1145/3589334.3645381},
  pages     = {2359--2370},
  publisher = {ACM},
  title     = {Towards explainable harmful meme detection through multimodal debate between large language models},
  url       = {https://doi.org/10.1145/3589334.3645381},
  year      = {2024},
}

@inproceedings{xu2024exploring,
  address   = {Bangkok, Thailand},
  author    = {Xu, Yanzhi and Hua, Yueying and Li, Shichen and Wang, Zhongqing},
  booktitle = {Proceedings of the 62nd Annual Meeting of the Association for Computational Linguistics (Volume 1: Long Papers)},
  series    = {ACL~'24},
  doi       = {10.18653/v1/2024.acl-long.6},
  pages     = {91--101},
  publisher = {Association for Computational Linguistics},
  title     = {Exploring chain-of-thought for multi-modal metaphor detection},
  url       = {https://aclanthology.org/2024.acl-long.6/},
  year      = {2024},
}

@inproceedings{lin-etal-2023-beneath,
  address   = {Singapore},
  author    = {Lin, Hongzhan and
               Luo, Ziyang and
               Ma, Jing and
               Chen, Long},
  editor    = {Bouamor, Houda and
               Pino, Juan and
               Bali, Kalika},
  booktitle = {Findings of the Association for Computational Linguistics: {EMNLP} 2023},
  series    = {Findings~'23},
  doi       = {10.18653/v1/2023.findings-emnlp.611},
  pages     = {9114--9128},
  publisher = {Association for Computational Linguistics},
  title     = {Beneath the surface: Unveiling harmful memes with multimodal reasoning distilled from large language models},
  url       = {https://aclanthology.org/2023.findings-emnlp.611},
  year      = {2023},
}

@article{kumari2024m3hop,
  author  = {Kumari, Gitanjali and Jain, Kirtan and Ekbal, Asif},
  journal = {ArXiv preprint},
  title   = {{M3Hop-CoT}: Misogynous meme identification with multimodal multi-hop chain-of-thought},
  url     = {https://arxiv.org/abs/2410.09220},
  volume  = {arXiv:2410.09220},
  year    = {2024},
}

@inproceedings{hasanain2024can,
  address   = {Torino, Italy},
  author    = {Hasanain, Maram and
               Ahmad, Fatema and
               Alam, Firoj},
  editor    = {Calzolari, Nicoletta and
               Kan, Min-Yen and
               Hoste, Veronique and
               Lenci, Alessandro and
               Sakti, Sakriani and
               Xue, Nianwen},
  booktitle = {Proceedings of the 2024 Joint International Conference on Computational Linguistics, Language Resources and Evaluation},
  series    = {LREC-COLING~'24},
  pages     = {2724--2744},
  publisher = {ELRA and ICCL},
  title     = {Can {GPT-4} identify propaganda? {A}nnotation and detection of propaganda spans in news articles},
  url       = {https://aclanthology.org/2024.lrec-main.244},
  year      = {2024},
}

@inproceedings{shridhar2023distilling,
  address   = {Toronto, Canada},
  author    = {Shridhar, Kumar and
               Stolfo, Alessandro and
               Sachan, Mrinmaya},
  editor    = {Rogers, Anna and
               Boyd-Graber, Jordan and
               Okazaki, Naoaki},
  booktitle = {Findings of the Association for Computational Linguistics: {ACL} 2023},
  series    = {Findings~'23},
  doi       = {10.18653/v1/2023.findings-acl.441},
  pages     = {7059--7073},
  publisher = {Association for Computational Linguistics},
  title     = {Distilling reasoning capabilities into smaller language models},
  url       = {https://aclanthology.org/2023.findings-acl.441},
  year      = {2023},
}

@article{zhu2024distilling,
  author    = {Zhu, Xunyu and Li, Jian and Liu, Yong and Ma, Can and Wang, Weiping},
  doi       = {10.1016/j.neunet.2024.106594},
  journal   = {Neural Networks},
  pages     = {106594},
  publisher = {Elsevier},
  title     = {Distilling mathematical reasoning capabilities into small language models},
  url       = {https://doi.org/10.1016/j.neunet.2024.106594},
  volume    = {179},
  year      = {2024},
}

@inproceedings{burbi2023mapping,
  address   = {Paris, France},
  author    = {Burbi, Giovanni and Baldrati, Alberto and Agnolucci, Lorenzo and Bertini, Marco and Del Bimbo, Alberto},
  booktitle = {Proceedings of the {IEEE/CVF} International Conference on Computer Vision Workshops},
  series    = {ICCVW~'23},
  doi       = {10.1109/ICCVW60793.2023.00303},
  pages     = {2832--2836},
  publisher = {IEEE},
  title     = {Mapping memes to words for multimodal hateful meme classification},
  url       = {https://doi.org/10.1109/ICCVW60793.2023.00303},
  year      = {2023},
}

@inproceedings{cao-etal-2022-prompting,
  address   = {Abu Dhabi, United Arab Emirates},
  author    = {Cao, Rui and
               Lee, Roy Ka-Wei and
               Chong, Wen-Haw and
               Jiang, Jing},
  editor    = {Goldberg, Yoav and
               Kozareva, Zornitsa and
               Zhang, Yue},
  booktitle = {Proceedings of the 2022 Conference on Empirical Methods in Natural Language Processing},
  series    = {EMNLP~'22},
  doi       = {10.18653/v1/2022.emnlp-main.22},
  pages     = {321--332},
  publisher = {Association for Computational Linguistics},
  title     = {Prompting for multimodal hateful meme classification},
  url       = {https://aclanthology.org/2022.emnlp-main.22},
  year      = {2022},
}

@article{wu2024multimodal,
  author    = {Wu, Fan and Chen, Guolian and Cao, Junkuo and Yan, Yuhan and Li, Zhongneng},
  doi       = {10.3390/electronics13142780},
  journal   = {Electronics},
  number    = {14},
  pages     = {2780},
  publisher = {MDPI},
  title     = {Multimodal hateful meme classification based on transfer learning and a cross-mask mechanism},
  url       = {https://doi.org/10.3390/electronics13142780},
  volume    = {13},
  year      = {2024},
}

@inproceedings{yang2024uncertainty,
  address   = {Bangkok, Thailand},
  author    = {Yang, Chuanpeng and Liu, Yaxin and Zhu, Fuqing and Han, Jizhong and Hu, Songlin},
  booktitle = {Proceedings of the 62nd Annual Meeting of the Association for Computational Linguistics (Volume 1: Long Papers)},
  series    = {ACL~'24},
  pages     = {4361--4371},
  publisher = {Association for Computational Linguistics},
  title     = {Uncertainty-guided modal rebalance for hateful memes detection},
  url       = {https://aclanthology.org/2024.acl-long.239/},
  year      = {2024},
}

@article{james1984estimating,
  author    = {James, Lawrence R and Demaree, Robert G and Wolf, Gerrit},
  doi       = {10.1037/0021-9010.69.1.85},
  journal   = {Journal of Applied Psychology},
  number    = {1},
  pages     = {85--98},
  publisher = {American Psychological Association},
  title     = {Estimating within-group interrater reliability with and without response bias},
  url       = {https://doi.org/10.1037/0021-9010.69.1.85},
  volume    = {69},
  year      = {1984},
}

@inproceedings{zhang2020bertscoreevaluatingtextgeneration,
  address   = {Addis Ababa, Ethiopia},
  author    = {Tianyi Zhang and
               Varsha Kishore and
               Felix Wu and
               Kilian Q. Weinberger and
               Yoav Artzi},
  booktitle = {Proceedings of the International Conference on Learning Representations},
  series    = {ICLR~'20},
  publisher = {OpenReview.net},
  title     = {{BERTScore}: Evaluating text generation with {BERT}},
  url       = {https://openreview.net/forum?id=SkeHuCVFDr},
  year      = {2020},
}

@inproceedings{banerjee-lavie-2005-meteor,
  address   = {Ann Arbor, MI, USA},
  author    = {Banerjee, Satanjeev and
               Lavie, Alon},
  editor    = {Goldstein, Jade and
               Lavie, Alon and
               Lin, Chin-Yew and
               Voss, Clare},
  booktitle = {Proceedings of the {ACL} Workshop on Intrinsic and Extrinsic Evaluation Measures for Machine Translation and/or Summarization},
  series    = {ACL~'05 Workshops},
  pages     = {65--72},
  publisher = {Association for Computational Linguistics},
  title     = {{METEOR}: An automatic metric for {MT} evaluation with improved correlation with human judgments},
  url       = {https://aclanthology.org/W05-0909},
  year      = {2005},
}

@inproceedings{luo2025semi,
  address   = {Albuquerque, NM, USA},
  author    = {Luo, Junyu and Luo, Xiao and Chen, Xiusi and Xiao, Zhiping and Ju, Wei and Zhang, Ming},
  booktitle = {Findings of the Association for Computational Linguistics: {NAACL} 2025},
  series    = {Findings~'25},
  pages     = {2795--2808},
  publisher = {Association for Computational Linguistics},
  title     = {Semi-supervised fine-tuning for large language models},
  url       = {https://aclanthology.org/2025.findings-naacl.151/},
  year      = {2025},
}

@article{shahroor2026memelens,
  title={MemeLens: Multilingual Multitask VLMs for Memes},
  author={Shahroor, Ali Ezzat and Kmainasi, Mohamed Bayan and Hasnat, Abul and Dimitrov, Dimitar and Martino, Giovanni Da San and Nakov, Preslav and Alam, Firoj},
  journal={arXiv preprint arXiv:2601.12539},
  year={2026}
}

@inproceedings{li2024self,
  address   = {Miami, FL, USA},
  author    = {Li, Jian and Huang, Haojing and Zhang, Yujia and Xu, Pengfei and Chen, Xi and Song, Rui and Shi, Lida and Wang, Jingwen and Xu, Hao},
  booktitle = {Findings of the Association for Computational Linguistics: {EMNLP} 2024},
  series    = {Findings~'24},
  pages     = {14452--14466},
  publisher = {Association for Computational Linguistics},
  title     = {Self-supervised preference optimization: Enhance your language model with preference degree awareness},
  url       = {https://aclanthology.org/2024.findings-emnlp.845/},
  year      = {2024},
}

@inproceedings{hasanain-etal-2025-propxplain,
  address   = {Suzhou, China},
  author    = {Hasanain, Maram and
               Hasan, Md Arid and
               Kmainasi, Mohamed Bayan and
               Sartori, Elisa and
               Shahroor, Ali Ezzat and
               {Da San Martino}, Giovanni and
               Alam, Firoj},
  editor    = {Christodoulopoulos, Christos and
               Chakraborty, Tanmoy and
               Rose, Carolyn and
               Peng, Violet},
  booktitle = {Findings of the Association for Computational Linguistics: {EMNLP} 2025},
  series    = {Findings~'25},
  doi       = {10.18653/v1/2025.findings-emnlp.1296},
  isbn      = {979-8-89176-335-7},
  pages     = {23855--23863},
  publisher = {Association for Computational Linguistics},
  title     = {{PropXplain}: Can {LLMs} enable explainable propaganda detection?},
  url       = {https://aclanthology.org/2025.findings-emnlp.1296/},
  year      = {2025},
}

@article{firooz2025scaling,
  author  = {Firooz, Hamed and Liu, Rui and Lu, Yuchen and Hou, Zhenyu and Xiong, Fangzhou and Zhang, Xiaoyang and Jian, Changshu and Zhu, Zhicheng and Ma, Jiayuan and Tao, Jacob and others},
  journal = {ArXiv preprint},
  title   = {Scaling reinforcement learning for content moderation with large language models},
  url     = {https://arxiv.org/abs/2512.20061},
  volume  = {arXiv:2512.20061},
  year    = {2025},
}

@article{yu2025dapo,
  author  = {Yu, Qiying and Zhang, Zheng and Zhu, Ruofei and Yuan, Yufeng and Zuo, Xiaochen and Yue, Yu and Dai, Weinan and Fan, Tiantian and Liu, Gaohong and Liu, Lingjun and others},
  journal = {ArXiv preprint},
  title   = {{DAPO}: An open-source {LLM} reinforcement learning system at scale},
  url     = {https://arxiv.org/abs/2503.14476},
  volume  = {arXiv:2503.14476},
  year    = {2025},
}

@article{abdin2024phi3technicalreporthighly,
  author  = {Abdin, Marah and Aneja, Jyoti and Awadalla, Hany and Awadallah, Ahmed and Awan, Ammar Ahmad and Bach, Nguyen and Bahree, Amit and Bakhtiari, Arash and Bao, Jianmin and Behl, Harkirat and others},
  journal = {ArXiv preprint},
  title   = {{Phi-3} technical report: A highly capable language model locally on your phone},
  url     = {https://arxiv.org/abs/2404.14219},
  volume  = {arXiv:2404.14219},
  year    = {2024},
}

@article{wang2025internvl35advancingopensourcemultimodal,
  author  = {Weiyun Wang and Zhangwei Gao and Lixin Gu and Hengjun Pu and Long Cui and Xingguang Wei and Zhaoyang Liu and Linglin Jing and Shenglong Ye and Jie Shao and Zhaokai Wang and Zhe Chen and Hongjie Zhang and Ganlin Yang and Haomin Wang and Qi Wei and Jinhui Yin and Wenhao Li and Erfei Cui and Guanzhou Chen and Zichen Ding and Changyao Tian and Zhenyu Wu and Jingjing Xie and Zehao Li and Bowen Yang and Yuchen Duan and Xuehui Wang and Zhi Hou and Haoran Hao and Tianyi Zhang and Songze Li and Xiangyu Zhao and Haodong Duan and Nianchen Deng and Bin Fu and Yinan He and Yi Wang and Conghui He and Botian Shi and Junjun He and Yingtong Xiong and Han Lv and Lijun Wu and Wenqi Shao and Kaipeng Zhang and Huipeng Deng and Biqing Qi and Jiaye Ge and Qipeng Guo and Wenwei Zhang and Songyang Zhang and Maosong Cao and Junyao Lin and Kexian Tang and Jianfei Gao and Haian Huang and Yuzhe Gu and Chengqi Lyu and Huanze Tang and Rui Wang and Haijun Lv and Wanli Ouyang and Limin Wang and Min Dou and Xizhou Zhu and Tong Lu and Dahua Lin and Jifeng Dai and Weijie Su and Bowen Zhou and Kai Chen and Yu Qiao and Wenhai Wang and Gen Luo},
  journal = {ArXiv preprint},
  title   = {{InternVL3.5}: Advancing open-source multimodal models in versatility, reasoning, and efficiency},
  url     = {https://arxiv.org/abs/2508.18265},
  volume  = {arXiv:2508.18265},
  year    = {2025},
}

@inproceedings{sharma2020semeval,
  address   = {Barcelona (online)},
  author    = {Sharma, Chhavi and
               Bhageria, Deepesh and
               Scott, William and
               PYKL, Srinivas and
               Das, Amitava and
               Chakraborty, Tanmoy and
               Pulabaigari, Viswanath and
               Gamb{\"a}ck, Bj{\"o}rn},
  editor    = {Herbelot, Aurelie and
               Zhu, Xiaodan and
               Palmer, Alexis and
               Schneider, Nathan and
               May, Jonathan and
               Shutova, Ekaterina},
  booktitle = {Proceedings of the Fourteenth Workshop on Semantic Evaluation},
  series    = {SemEval~'20},
  doi       = {10.18653/v1/2020.semeval-1.99},
  pages     = {759--773},
  publisher = {Association for Computational Linguistics},
  title     = {{SemEval}-2020 Task 8: Memotion analysis --- the visuo-lingual metaphor!},
  url       = {https://aclanthology.org/2020.semeval-1.99},
  year      = {2020},
}

@article{wang2026v,
  author  = {Wang, Han and Yang, Yi and Hu, Jingyuan and Zhu, Minfeng and Chen, Wei},
  journal = {ArXiv preprint},
  title   = {{V-Zero}: Self-improving multimodal reasoning with zero annotation},
  url     = {https://arxiv.org/abs/2601.10094},
  volume  = {arXiv:2601.10094},
  year    = {2026},
}

@inproceedings{kmainasi2024native,
  address   = {Doha, Qatar},
  author    = {Kmainasi, Mohamed Bayan and Khan, Rakif and Shahroor, Ali Ezzat and Bendou, Boushra and Hasanain, Maram and Alam, Firoj},
  booktitle = {Proceedings of the International Conference on Web Information Systems Engineering},
  series    = {WISE~'24},
  doi       = {10.1007/978-981-96-0576-7_30},
  pages     = {406--420},
  publisher = {Springer},
  title     = {Native vs non-native language prompting: A comparative analysis},
  url       = {https://doi.org/10.1007/978-981-96-0576-7_30},
  year      = {2024},
}

@inproceedings{kmainasi2026thinking,
  address   = {Dubai, United Arab Emirates},
  author    = {Kmainasi, Mohamed Bayan and Kutlu, Mucahid and Shahroor, Ali Ezzat and Hasnat, Abul and Alam, Firoj},
  booktitle = {Companion Proceedings of the ACM Web Conference},
  series    = {WWW Companion~'26},
  doi       = {10.1145/3774905.3795465},
  pages     = {1--10},
  publisher = {ACM},
  title     = {Can thinking models think to detect hateful memes?},
  url       = {https://doi.org/10.1145/3774905.3795465},
  year      = {2026},
}

@inproceedings{bao2022beit,
  address   = {Virtual},
  author    = {Bao, Hangbo and Dong, Li and Piao, Songhao and Wei, Furu},
  booktitle = {Proceedings of the International Conference on Learning Representations},
  series    = {ICLR~'22},
  publisher = {OpenReview.net},
  title     = {{BEiT}: {BERT} pre-training of image transformers},
  url       = {https://openreview.net/forum?id=p-BhZSz59o4},
  year      = {2022},
}

@inproceedings{liu2022convnet,
  address   = {New Orleans, LA, USA},
  author    = {Liu, Zhuang and Mao, Hanzi and Wu, Chao-Yuan and Feichtenhofer, Christoph and Darrell, Trevor and Xie, Saining},
  booktitle = {Proceedings of the {IEEE/CVF} Conference on Computer Vision and Pattern Recognition},
  series    = {CVPR~'22},
  doi       = {10.1109/CVPR52688.2022.01167},
  pages     = {11966--11976},
  publisher = {IEEE},
  title     = {A {ConvNet} for the 2020s},
  url       = {https://doi.org/10.1109/CVPR52688.2022.01167},
  year      = {2022},
}

@article{oquab2024dinov2,
  author  = {Oquab, Maxime and Darcet, Timoth{\'e}e and Moutakanni, Th{\'e}o and Vo, Huy and Szafraniec, Marc and Khalidov, Vasil and Fernandez, Pierre and Haziza, Daniel and Massa, Francisco and El-Nouby, Alaaeldin and Assran, Mahmoud and Ballas, Nicolas and Galuba, Wojciech and Howes, Russell and Huang, Po-Yao and Li, Shang-Wen and Misra, Ishan and Rabbat, Michael and Sharma, Vasu and Synnaeve, Gabriel and Xu, Hu and J{\'e}gou, Herv{\'e} and Mairal, Julien and Labatut, Patrick and Joulin, Armand and Bojanowski, Piotr},
  journal = {Transactions on Machine Learning Research},
  title   = {{DINOv2}: Learning robust visual features without supervision},
  url     = {https://openreview.net/forum?id=a68SUt6zFt},
  year    = {2024},
}

@inproceedings{he2016resnet,
  address   = {Las Vegas, NV, USA},
  author    = {He, Kaiming and Zhang, Xiangyu and Ren, Shaoqing and Sun, Jian},
  booktitle = {Proceedings of the {IEEE} Conference on Computer Vision and Pattern Recognition},
  series    = {CVPR~'16},
  doi       = {10.1109/CVPR.2016.90},
  pages     = {770--778},
  publisher = {IEEE},
  title     = {Deep residual learning for image recognition},
  url       = {https://doi.org/10.1109/CVPR.2016.90},
  year      = {2016},
}

@inproceedings{liu2021swin,
  address   = {Montreal, QC, Canada},
  author    = {Liu, Ze and Lin, Yutong and Cao, Yue and Hu, Han and Wei, Yixuan and Zhang, Zheng and Lin, Stephen and Guo, Baining},
  booktitle = {Proceedings of the {IEEE/CVF} International Conference on Computer Vision},
  series    = {ICCV~'21},
  doi       = {10.1109/ICCV48922.2021.00986},
  pages     = {9992--10002},
  publisher = {IEEE},
  title     = {{Swin Transformer}: Hierarchical vision transformer using shifted windows},
  url       = {https://doi.org/10.1109/ICCV48922.2021.00986},
  year      = {2021},
}

@inproceedings{dosovitskiy2021vit,
  address   = {Virtual},
  author    = {Dosovitskiy, Alexey and Beyer, Lucas and Kolesnikov, Alexander and Weissenborn, Dirk and Zhai, Xiaohua and Unterthiner, Thomas and Dehghani, Mostafa and Minderer, Matthias and Heigold, Georg and Gelly, Sylvain and Uszkoreit, Jakob and Houlsby, Neil},
  booktitle = {Proceedings of the International Conference on Learning Representations},
  series    = {ICLR~'21},
  publisher = {OpenReview.net},
  title     = {An image is worth {16x16} words: Transformers for image recognition at scale},
  url       = {https://openreview.net/forum?id=YicbFdNTTy},
  year      = {2021},
}

@inproceedings{devlin2019bert,
  address   = {Minneapolis, MN, USA},
  author    = {Devlin, Jacob and
               Chang, Ming-Wei and
               Lee, Kenton and
               Toutanova, Kristina},
  editor    = {Burstein, Jill and
               Doran, Christy and
               Solorio, Thamar},
  booktitle = {Proceedings of the 2019 Conference of the North {A}merican Chapter of the Association for Computational Linguistics: Human Language Technologies, Volume 1 (Long and Short Papers)},
  series    = {NAACL~'19},
  doi       = {10.18653/v1/N19-1423},
  pages     = {4171--4186},
  publisher = {Association for Computational Linguistics},
  title     = {{BERT}: Pre-training of deep bidirectional transformers for language understanding},
  url       = {https://aclanthology.org/N19-1423/},
  year      = {2019},
}

@article{abdelali2021qarib,
  author  = {Abdelali, Ahmed and Hassan, Sabit and Mubarak, Hamdy and Darwish, Kareem and Samih, Younes},
  journal = {ArXiv preprint},
  title   = {Pre-training {BERT} on {A}rabic tweets: Practical considerations},
  url     = {https://arxiv.org/abs/2102.10684},
  volume  = {arXiv:2102.10684},
  year    = {2021},
}

@inproceedings{antoun2020arabert,
  address   = {Marseille, France},
  author    = {Antoun, Wissam and
               Baly, Fady and
               Hajj, Hazem},
  editor    = {Al-Khalifa, Hend and
               Magdy, Walid and
               Darwish, Kareem and
               Elsayed, Tamer and
               Mubarak, Hamdy},
  booktitle = {Proceedings of the 4th Workshop on Open-Source {A}rabic Corpora and Processing Tools, with a Shared Task on Offensive Language Detection},
  series    = {OSACT~'20},
  pages     = {9--15},
  publisher = {European Language Resource Association},
  title     = {{AraBERT}: Transformer-based model for {A}rabic language understanding},
  url       = {https://aclanthology.org/2020.osact-1.2/},
  year      = {2020},
}

@article{sanh2019distilbert,
  author  = {Sanh, Victor and Debut, Lysandre and Chaumond, Julien and Wolf, Thomas},
  journal = {ArXiv preprint},
  title   = {{DistilBERT}, a distilled version of {BERT}: Smaller, faster, cheaper and lighter},
  url     = {https://arxiv.org/abs/1910.01108},
  volume  = {arXiv:1910.01108},
  year    = {2019},
}

@inproceedings{conneau2020xlmr,
  address   = {Online},
  author    = {Conneau, Alexis and
               Khandelwal, Kartikay and
               Goyal, Naman and
               Chaudhary, Vishrav and
               Wenzek, Guillaume and
               Guzm{\'a}n, Francisco and
               Grave, Edouard and
               Ott, Myle and
               Zettlemoyer, Luke and
               Stoyanov, Veselin},
  editor    = {Jurafsky, Dan and
               Chai, Joyce and
               Schluter, Natalie and
               Tetreault, Joel},
  booktitle = {Proceedings of the 58th Annual Meeting of the Association for Computational Linguistics},
  series    = {ACL~'20},
  doi       = {10.18653/v1/2020.acl-main.747},
  pages     = {8440--8451},
  publisher = {Association for Computational Linguistics},
  title     = {Unsupervised cross-lingual representation learning at scale},
  url       = {https://aclanthology.org/2020.acl-main.747/},
  year      = {2020},
}

@inproceedings{skalse2022defining,
  address   = {New Orleans, LA, USA},
  author    = {Skalse, Joar and Howe, Nikolaus H. R. and Krasheninnikov, Dmitrii and Krueger, David},
  booktitle = {Proceedings of the 36th International Conference on Neural Information Processing Systems},
  series    = {NeurIPS~'22},
  title     = {Defining and characterizing reward hacking},
  year      = {2022},
}

@inproceedings{singhal2024long,
  address   = {Philadelphia, PA, USA},
  author    = {Singhal, Prasann and Goyal, Tanya and Xu, Jiacheng and Durrett, Greg},
  booktitle = {Proceedings of the First Conference on Language Modeling},
  series    = {COLM~'24},
  title     = {A long way to go: Investigating length correlations in {RLHF}},
  url       = {https://openreview.net/forum?id=G8LaO1P0xv},
  year      = {2024},
}

@article{fanarteam2026fanar20arabicgenerative,
  author  = {{FANAR TEAM} and Abbas, Ummar and Ahmad, Mohammad Shahmeer and Ahmad, Minhaj and Al-Homaid, Abdulaziz and Al-Nuaimi, Anas and Altinisik, Enes and Asgari, Ehsaneddin and Chawla, Sanjay and Chowdhury, Shammur and Dalvi, Fahim and Darwish, Kareem and Durrani, Nadir and Elfeky, Mohamed and Elmagarmid, Ahmed and Eltabakh, Mohamed and Ersoy, Asim and Fatehkia, Masoomali and Hashim, Mohammed Qusay and Hawasly, Majd and Hefeeda, Mohamed and Husaini, Mus'ab and Isufaj, Keivin and Jung, Soon-Gyo and Lachemat, Houssam and Lucas, Ji Kim and Mohamed, Abubakr and Mohiuddin, Tasnim and Mousi, Basel and Mubarak, Hamdy and Musleh, Ahmad and Ouzzani, Mourad and Sadeghi, Amin and Sencar, Husrev Taha and Shinoy, Mohammed and Sinan, Omar and Zhang, Yifan},
  journal = {ArXiv preprint},
  title   = {{Fanar}~2.0: {A}rabic generative {AI} stack},
  url     = {https://arxiv.org/abs/2603.16397},
  volume  = {arXiv:2603.16397},
  year    = {2026},
}

@inproceedings{dror2018hitchhiker,
  address   = {Melbourne, Australia},
  author    = {Dror, Rotem and Baumer, Gili and Shlomov, Segev and Reichart, Roi},
  booktitle = {Proceedings of the 56th Annual Meeting of the Association for Computational Linguistics (Volume 1: Long Papers)},
  series    = {ACL~'18},
  pages     = {1383--1392},
  publisher = {Association for Computational Linguistics},
  title     = {The hitchhiker's guide to testing statistical significance in natural language processing},
  year      = {2018},
}

@article{holm1979simple,
  author    = {Holm, Sture},
  journal   = {Scandinavian Journal of Statistics},
  pages     = {65--70},
  publisher = {JSTOR},
  title     = {A simple sequentially rejective multiple test procedure},
  volume    = {6},
  year      = {1979},
}

@article{gwet2008computing,
  title={Computing inter-rater reliability and its variance in the presence of high agreement},
  author={Gwet, Kilem Li},
  journal={British Journal of Mathematical and Statistical Psychology},
  volume={61},
  number={1},
  pages={29--48},
  year={2008},
  publisher={Wiley Online Library}
}

@article{ren2026survey,
  title={A survey of multimodal hate meme detection},
  author={Ren, Chengjuan and Jeong, Dongwon and Wu, Ming and Huang, Yi and Gao, Yuhan and Li, Yuejia},
  journal={Expert Systems with Applications},
  pages={132507},
  year={2026},
  publisher={Elsevier}
}

\end{document}